\documentclass[lettersize,journal]{IEEEtran}
\bstctlcite{bstctl:etal, bstctl:nodash, bstctl:simpurl}

\usepackage{amsmath}
\usepackage{adjustbox}
\usepackage{amssymb}
\usepackage{booktabs}
\usepackage{xspace}
\usepackage[table]{xcolor}
\usepackage{graphicx}
\usepackage{multirow}
\usepackage{pifont}
\usepackage{color}
\usepackage{cite}
\usepackage{bm}
\usepackage[figureposition=bottom,tableposition=top]{caption}

\definecolor{gray}{RGB}{142,142,142}
\definecolor{gray9}{gray}{.9}
\definecolor{gray95}{gray}{.95}
\definecolor{gray8}{gray}{.8}
\definecolor{gray85}{gray}{.85}
\definecolor{darkgreen}{RGB}{0, 150, 0}
\definecolor{darkred}{RGB}{200, 0, 0}

\makeatletter
\DeclareRobustCommand\onedot{\futurelet\@let@token\@onedot}
\def\@onedot{\ifx\@let@token.\else.\null\fi\xspace}

\makeatother

\usepackage[colorlinks,pagebackref=false,citecolor=blue,bookmarks=false,hypertexnames=true]{hyperref}

\title{\LARGE \bf Neuromorphic LiDAR-based Bird's Eye View Object Detection using Energy-efficient Spiking Neural Networks}

\author{Sambit Mohapatra$^{1,3}$, 
Senthil Yogamani$^{2}$, 
Heinrich Gotzig$^{1}$ and
Patrick M\"ader$^{3}$\quad
\\

{$^{1}$Valeo, Germany\quad
$^{2}$Valeo, Ireland\quad
$^{3}$TU Ilmenau, Germany}
}

\begin{document}
\maketitle
\begin{abstract}

Autonomous driving perception demands accurate and efficient processing
of three-dimensional sensor data under strict power constraints.
Traditional convolutional neural networks achieve strong detection
accuracy but are computationally intensive, limiting their suitability
for deployment on resource-constrained neuromorphic platforms. Spiking
neural networks offer a compelling alternative through event-driven
sparse computation, yet their application to complex real-world
perception tasks such as three-dimensional object detection remains
limited. In this work, we propose an end-to-end spiking encoder-decoder
network for object detection in bird's eye view representations of
LiDAR point clouds, trained using surrogate gradient backpropagation.
We train two variants: a membrane potential variant that reads
continuous neuron state at the output stage for maximum accuracy,
achieving $92.05$/$87.04$/$86.51$ AP at $\mathrm{IoU}\!=\!0.5$
(Easy/Moderate/Hard), and, a fully binary spiking variant that operates
exclusively on spike trains at every layer for direct neuromorphic
deployment.
We evaluate four input spike encoding strategies and demonstrate
that allowing the network to learn spike representations directly
from data outperforms hand-crafted Poisson, latency, and z-axis
encoding schemes on the KITTI benchmark, where sequential frames
are unavailable and the BEV input is presented repeatedly across
timesteps as a proxy for temporal streaming. A block-wise energy analysis demonstrates a $3.33\times$ reduction in synaptic operation energy over an equivalent CNN under
conservative loop-based operation. Together, these results demonstrate the viability of spiking neural networks for accurate and energy-efficient neuromorphic perception in autonomous driving.


\end{abstract}
\section{Introduction}
Object detection is a fundamental task in autonomous
driving and advanced driver assistance systems, with direct impact
on downstream functions including collision avoidance, path
planning, and driving decision making \cite{joseph2021autonomous}. LiDAR sensors have become
central to state-of-the-art three-dimensional object detection
pipelines, providing dense and accurate spatial measurements of
the environment that enable robust perception under varying
lighting and atmospheric conditions compared to depth from cameras \cite{kumar2018near}.

Convolutional neural networks~(CNNs) have been the dominant
paradigm for processing LiDAR point clouds. Three-dimensional CNN
architectures operate directly on voxelized point clouds using 3D
convolutions, while 2D approaches project the point
cloud into compact representations such as bird's eye view~(BEV)
images or range images, enabling faster inference through 2D
convolutions. Despite their accuracy, CNNs process every spatial
location in the sensor field of view with equal computational
weight. For
sparse inputs such as LiDAR BEV maps---where the majority of
cells contain no returns---this leads to significant wasted
computation and energy, processing empty cells with the same cost
as informative measurements. Furthermore, the inherently
frame-based operating principle of CNNs does not differentiate
between information-rich regions and redundant background, nor
does it exploit temporal persistence across consecutive sensor
frames.

Spiking neural networks~(SNNs) offer a principled solution to
these limitations by drawing inspiration from biological neural
computation. Information is represented and communicated as
discrete binary spikes, replacing the multiply-accumulate~(MAC)
operations of CNNs with conditional accumulate-only~(AC)
operations that are triggered exclusively by spike events. Since
inactive neurons contribute no synaptic operations, the
computational cost scales directly with the firing sparsity of
the network rather than its total capacity---a property that is
particularly advantageous for sparse inputs such as LiDAR BEV
maps. Beyond computational sparsity, spiking neurons maintain
internal state through their membrane potential, which integrates
incoming spikes across multiple timesteps. This
temporal integration allows the network to accumulate evidence
progressively, suppressing transient noise and reinforcing
persistent spatial features. On
neuromorphic hardware platforms such as Intel Loihi~2~\cite{orchard2021loihi},
IBM TrueNorth~\cite{merolla2014million}, and
SpiNNaker~\cite{furber2014spinnaker}, these properties translate
directly into measurable energy savings. The replacement of MAC
with AC operations alone reduces per-synaptic-event energy by a
factor of approximately $5\times$ at 45\,nm process
technology~\cite{horowitz20141}, and this saving compounds
multiplicatively with network sparsity.

These properties become increasingly relevant as automotive
sensor technology advances toward higher frame rates and
event-driven modalities. Next-generation solid-state LiDAR
sensors are targeting scan rates well beyond the 10--20\,Hz of
current rotating systems~\cite{li2020lidar}, 
and event cameras operating at megahertz temporal resolution
are beginning to appear in automotive perception
research~\cite{gallego2020event}.Radars are sparser than LiDARs and CNNs are less efficient \cite{schramm2024bevcar}. 
For CNN-based pipelines, higher frame rates translate directly
to proportionally higher computational and energy costs, since
every frame requires a complete forward pass at full precision
regardless of how much the scene has changed between frames.
SNNs are architecturally well-suited to this trend: the
timesteps map naturally onto the sensor frame rate,
each new measurement advancing the membrane potential
integration by one step rather than triggering a complete
recomputation from scratch. 
This temporal and energetic scalability positions SNNs
as a particularly promising architecture for the high-rate,
event-driven sensing pipelines that next-generation autonomous
vehicles will rely upon. While the experiments in this work
operate on pre-recorded static frames from an established
benchmark, the architecture is designed with sequential
deployment in mind: the temporal window maps directly
onto a live sensor stream in a deployed system, where each
incoming frame naturally advances the spiking dynamics without
any explicit frame repetition.

Despite these theoretical advantages, SNNs remain predominantly
applied to classification tasks, with only a small number of
published works addressing complex real-world perception tasks
such as 3D object detection, which requires multi-scale feature
extraction, simultaneous regression and classification outputs,
and operation on large high-resolution spatial maps. This gap is
significant: the perception tasks that would benefit most from
the energy efficiency of SNNs are precisely those that have
proven most challenging to port from CNN architectures.


This paper addresses this gap directly. We propose an end-to-end
spiking encoder-decoder network for object detection in BEV of LiDAR point clouds from the KITTI benchmark~\cite{geiger2013vision}, trained using surrogate
gradient backpropagation implemented in the \texttt{snntorch}
library~\cite{eshraghian2023training}. We demonstrate that spiking neural networks can match the detection accuracy of full-precision CNNs while consuming a
fraction of their inference energy, and that the
accuracy--efficiency trade-off historically associated with
SNN-based perception can be substantially reduced through
careful architecture design, spike-domain loss functions, and
training configuration.

The main contributions of this work are as follows:

\begin{itemize}

    \item We propose a spiking encoder-decoder architecture for
    bird's eye view LiDAR object detection that integrates
    spatial feature extraction with the temporal dynamics of
    leaky integrate-and-fire neurons, enabling low-power
    event-driven processing of sparse BEV maps.

    \item We present solutions to the key challenges of applying
    SNNs to complex perception tasks, including end-to-end
    surrogate gradient training and two novel spike-domain loss
    functions: (i)~a \textit{two-point temporal BCE and Dice
    loss} for keypoint detection that supervises both early and
    full temporal windows of spike activity to enforce spatial
    coherence of object-center predictions, and (ii)~a
    \textit{population-coded spike regression loss} for bounding
    box estimation that encodes continuous physical dimensions
    through the collective firing rate of a local $k \times k$
    neuron population---eliminating the need to discretise
    regression targets into bins. Rotation prediction is
    formulated as a spike-compatible classification task over
    discretised orientation bins.

    \item We train and evaluate two inference variants---membrane
    potential and fully binary spiking---demonstrating that the
    fully spiking variant achieves competitive detection accuracy
    with no continuous-valued activations at any stage.

    \item We conduct a comprehensive block-wise synaptic energy
    analysis by mapping layer-wise firing rates to a 45\,nm
    technology energy model, quantifying a $3.33\times$ inference
    energy reduction over an equivalent CNN under conservative
    loop-based simulation---rising to an estimated $43\times$ on
    dedicated neuromorphic hardware---and identifying the
    accumulate-only operation substitution and $88.19\%$ firing
    sparsity as the two independent mechanisms driving this gain.

\end{itemize}


\section{Related Work}
\label{sec:related}

\subsection{LiDAR-Based 3D Object Detection}
\label{subsec:rw_detection}

LiDAR-based 3D object detection has been dominated by CNN
architectures operating on voxelized or projected
representations of point clouds. Early volumetric methods such
as Vote3Deep~\cite{engelcke2017vote3deep} and
VoxelNet~\cite{zhou2018voxelnet} apply sparse 3D convolutions
on voxelized point clouds, while
SECOND~\cite{yan2018second} accelerates this pipeline through
improved sparse convolution algorithms. Although accurate, 3D
convolutional architectures are computationally expensive and
scale poorly with field-of-view, driving interest in more
efficient 2D projection-based approaches.

Two-stage methods such as PointRCNN~\cite{shi2019pointrcnn}
generate region proposals from the raw point cloud using
PointNet++~\cite{qi2017pointnet++} and refine them in a second
stage, achieving strong accuracy at the cost of latency.
PV-RCNN~\cite{shi2020pv} further improves precision by fusing
point-level and voxel-level features into keypoint
representations, but the two-stage design remains a bottleneck
for real-time deployment.

Bird's eye view~(BEV) representations offer a practical
alternative: they preserve metric scale and aspect ratios,
naturally accommodate dense point clouds without increasing
map resolution, and enable efficient 2D convolutions.
PointPillars~\cite{lang2019pointpillars} encodes point clouds
into vertical pillars using PointNet~\cite{qi2017pointnet},
learning dense features over sparse BEV grids and remains a
strong real-time baseline. PIXOR~\cite{yang2018pixor} introduced
an anchor-free BEV detector using a BEV occupancy
map, demonstrating that simple 2D convolutional backbones can
achieve real-time 3D detection without anchor regression
overhead. BEVDetNet~\cite{mohapatra2021bevdetnet}
extends this with anchor-free keypoint-based object center
detection, achieving low inference latency with competitive
accuracy on KITTI~\cite{geiger2013vision}.
More recently, Naich et al.~\cite{naich2024lidar} proposed
enhanced voxel feature construction via intensity histograms,
though the 3D backbone limits real-time suitability.
Lis et al.~\cite{lis2025lift} introduced a fully quantized
INT8 pipeline targeting FPGA deployment, demonstrating that
hardware-aware design constraints can yield significant
efficiency gains without sacrificing accuracy.

Despite these advances, all of these methods operate on
dense floating-point activations and perform a full forward
pass for every input frame regardless of scene content. 
This frame-based, activation-dense processing is fundamentally
mismatched to the inherent sparsity of LiDAR data, where the
majority of the BEV map contains no returns.
These inefficiencies motivate the application of spiking neural
networks, which compute only where spike events occur.

\subsection{Spiking Neural Networks for Perception}
\label{subsec:rw_snn}

\subsubsection{\textbf{ANN-to-SNN Conversion}}

Early SNN perception methods circumvented the difficulty of
direct training by converting pre-trained CNNs to SNNs.
Cao et al.~\cite{cao2015spiking} and
Diehl et al.~\cite{diehl2015fast} demonstrated that CNNs
trained on simple datasets could be converted to functional
SNNs, but required large numbers of timesteps to approximate
the continuous activations of the source network, making
real-time deployment impractical.
Rueckauer et al.~\cite{rueckauer2017conversion} broadened the
conversion repertoire by proposing spiking equivalents of
pooling, batch normalization, and other standard CNN blocks,
enabling conversion of complex architectures trained on
ImageNet~\cite{deng2009imagenet} and
CIFAR-10~\cite{krizhevsky2009learning}.
Kim et al.~\cite{kim2020spiking} applied conversion to
image-based object detection on a challenging dataset~\cite{uricar2019challenges},
introducing signed neurons and channel-wise normalization to
handle the signed activations that detection requires.
More recently, Qu et al.~\cite{qu2024spiking} substantially
reduced the timestep requirement of conversion-based detectors
through time-varying spike thresholds, narrowing the accuracy
gap at practical latencies.

For 3D point cloud perception, SpikiLi~\cite{mohapatra2022spikili}, 
our prior work, applied conversion to BEV-based LiDAR object detection, demonstrating the
feasibility of SNN inference on point clouds. However, as
with other conversion methods, accuracy is bounded by the
pre-trained CNN weights and the approximation errors
introduced during conversion.
Tao et al.~\cite{tao2024spiking} extended conversion to point
cloud classification using spike-based
X-convolutions~\cite{li2018pointcnn}.
Lan et al.~\cite{lan2023efficient} improved conversion
fidelity through KL-divergence-based threshold initialization
to align spike activations with ReLU outputs, achieving
nearly lossless conversion for classification tasks. 
A fundamental limitation of all conversion-based
approaches is that they inherit the architectural constraints
of the source CNN and cannot exploit the temporal integration
properties of spiking neurons during training, limiting both
accuracy and energy efficiency.

\subsubsection{\textbf{Directly Trained SNNs}}

Direct training of SNNs using surrogate
gradients~\cite{neftci2019surrogate} and spatio-temporal
backpropagation~\cite{wu2018spatio} has produced significantly
stronger results than conversion, since the network is
optimized end-to-end for spike-based computation. This allows
the network to learn representations that are inherently
sparse and temporally structured rather than approximating a
pre-trained dense network.

For 2D object detection,
Li et al.~\cite{li2025brain} proposed multi-scale spiking
feature fusion achieving a strong energy-efficient benchmark,
while Cordone et al.~\cite{cordone2022detection} demonstrated
directly trained SNNs reaching competitive detection
performance on vehicular image datasets with up to 85\%
energy savings over equivalent CNNs.
Zhu et al.~\cite{zhu2024spiking} presented Spiking
Autonomous Driving (SAD), the first unified
end-to-end SNN addressing perception, prediction, and
planning for autonomous driving, underscoring the growing
maturity of directly trained SNNs for complex driving tasks.

For 3D point cloud perception, Ren et al.~\cite{ren2024spikingpointnet}
introduced Spiking PointNet, the first spiking model for
deep learning directly on point clouds, trained with a single
timestep but achieving improved performance with multiple
timesteps at inference. Wu et al.~\cite{wu2024pointtospike}
proposed point-to-spike residual learning for 3D point cloud
classification, designing spatial-aware kernel point spiking
neurons that relate spike generation to 3D point position and
stacking them into residual blocks for energy-efficient
feature learning.

For 3D detection,
SpikeCloudNet~\cite{ren2022spiking} applies direct training
to point cloud detection and reports competitive accuracy on
KITTI, representing the strongest directly-trained SNN
baseline for this task. Lian et al.~\cite{lian2020deep}
proposed DeepSCNN, integrating a spiking convolutional
network with temporal coding into the YOLOv2 architecture
and developing a novel preprocessing layer to translate 3D
point cloud data into spike time data, demonstrating SNN
applicability to real-time LiDAR detection.
However, none of these works address the challenge of
performing continuous regression directly from binary spike
outputs for 3D bounding box estimation---instead relying on
membrane potential readout or discretization schemes that
either require continuous-valued activations or sacrifice
precision. Furthermore, energy efficiency claims in prior
work are typically made at the network level without a
block-wise synaptic operation analysis that traces the
origin of savings to individual architectural stages.

Our work addresses these gaps by proposing a fully
end-to-end spiking network for LiDAR BEV detection trained
from scratch. We introduce novel spike-domain loss functions
for keypoint detection and population-coded bounding box
regression that operate directly on spike rates. Furthermore,
we provide a comprehensive block-wise energy analysis that
quantifies the efficiency contribution of each individual 
encoder and decoder stage. Furthermore, we provide a comprehensive block-wise energy analysis quantifying a $3.33\times$ reduction in synaptic operation energy 
over an equivalent CNN under conservative simulation conditions, 
rising to an estimated $43\times$ on dedicated neuromorphic 
hardware, with savings traced to individual encoder and decoder 
stages.
\section{Background}
\label{sec:background}

Neuromorphic computing draws inspiration from the signal
processing principles of biological neural systems to achieve
ultra-low-power event-driven computation. Spiking neural
networks~(SNNs) form the algorithmic foundation of neuromorphic
computing, communicating information via discrete binary
spikes---analogous to the voltage pulses of biological
neurons---rather than the continuous activations of conventional
networks.

\subsection{Spiking Neuron Models}
\label{subsec:neuron_models}

Unlike conventional neural networks that process inputs in a
single forward pass, spiking neural networks operate over
discrete timesteps, with neurons integrating incoming spikes
across time and emitting output spikes when their membrane
potential exceeds a threshold. Synaptic weights play the role of 
biological synapses, scaling the contribution of each
presynaptic spike to the postsynaptic neuron's membrane potential. Several neuron models exist, trading off biological fidelity against computational tractability.

\paragraph{\textbf{Integrate-and-Fire Model}}
The simplest spiking neuron model integrates incoming weighted
inputs into a membrane potential $V(t)$ and emits a spike when
$V(t)$ exceeds a threshold $V_{\mathrm{thr}}$, after which the
potential is reset. The model has no leak term, meaning the
membrane retains all accumulated charge indefinitely in the
absence of a spike. While computationally simple, this makes
the neuron sensitive to input history over arbitrarily long
windows, which can destabilise training on long sequences.

\paragraph{\textbf{Leaky Integrate-and-Fire Model}}
The Leaky Integrate-and-Fire~(LIF) model is the most widely
adopted spiking neuron model for deep SNN training due to its
balance between biological plausibility and computational
tractability~\cite{maass1997networks}, and is the neuron model
adopted throughout this work.
Biologically, a neuron maintains a voltage difference across its cell membrane, known 
as the membrane potential $V(t)$, which fluctuates as it 
integrates electrical charges from incoming spikes.
In continuous time, the membrane dynamics are governed by:

\begin{equation}
    \tau_m \frac{dV(t)}{dt}
    = -(V(t) - V_{\mathrm{rest}}) + R\,I(t),
    \label{eq:lif_continuous}
\end{equation}

\noindent where $\tau_m = RC$ is the membrane time constant,
$R$ the membrane resistance, $C$ the membrane capacitance,
$V_{\mathrm{rest}}$ the resting potential, and $I(t)$ the
total synaptic input current. The
leakage term $-(V(t)-V_{\mathrm{rest}})$ decays the membrane
potential toward rest in the absence of input, conferring
temporal memory that is absent in stateless CNN activations.

Setting $V_{\mathrm{rest}}\!=\!0$ and discretizing for
software simulation, the update equation implemented in
\texttt{snntorch}~\cite{eshraghian2023training} is:

\begin{equation}
    U[t+1] = \beta\,U[t] + W\,X[t+1] - S[t]\,U_{\mathrm{thr}},
    \label{eq:lif_discrete}
\end{equation}

\noindent where $U[t]$ is the membrane potential at timestep
$t$, $\beta \in (0,1)$ is the learnable decay factor controlling
the rate of membrane leakage, $W\,X[t+1]$ is the weighted
presynaptic spike input, and $-S[t]\,U_{\mathrm{thr}}$ implements
a subtract-and-continue reset: when a spike $S[t]\!=\!1$ is
emitted (i.e.\ $U[t] \geq U_{\mathrm{thr}}$), the threshold
voltage is subtracted from the membrane potential rather than
resetting it to zero, allowing any suprathreshold charge to
carry forward into the next timestep. A spike is generated
according to the Heaviside step function:

\begin{equation}
    S[t] = \Theta(U[t] - U_{\mathrm{thr}}),
    \label{eq:spike_fn}
\end{equation}

\noindent where $U_{\mathrm{thr}}$ is the learnable spike
threshold. Both $\beta$ and $U_{\mathrm{thr}}$ are optimized
end-to-end during training in our implementation.

\subsection{Spike Coding}
\label{subsec:spike_coding}

Spike coding defines how continuous-valued sensor data are
converted into binary spike trains for SNN processing.

\paragraph{\textbf{Rate coding}} encodes information in the spike count
over a time window. For a normalized input $x \in [0,1]$,
Poisson encoding~\cite{adrian1926impulses} generates a spike at
each timestep with probability:

\begin{equation}
    P(S[t] = 1) = x.
    \label{eq:rate_code}
\end{equation}

\noindent While robust, rate coding requires large $T$ to
represent high-precision values accurately, increasing latency
and energy cost~\cite{maass1997networks}.

\paragraph{\textbf{Temporal coding}} encodes information in spike timing
rather than count. In time-to-first-spike~(TTFS)
encoding~\cite{thorpe1996rapid}, larger inputs produce earlier
spikes. This is highly energy-efficient but yields very sparse
inputs that are difficult to train with gradient-based methods
and are sensitive to noise.

\paragraph{\textbf{Direct coding}} avoids fixed probabilistic or temporal
schemes by passing continuous inputs directly as injection
current to the first spiking layer, allowing the network to
learn an optimal task-specific encoding from
data~\cite{su2023deep}. This approach preserves input precision
and consistently outperforms hand-crafted coding schemes for
complex tasks. In this work we adopt a self-coding variant of
direct coding, in which the same BEV frame is presented
identically at each of the $T$ timesteps, allowing LIF neurons
to integrate evidence across repeated presentations and converge
to a stable internal representation. As confirmed by our
ablation in Table~\ref{tab:input-coding}, this outperforms
Poisson, latency, and z-axis encoding schemes for BEV-based
detection.

\subsection{Training with Surrogate Gradients}
\label{subsec:surrogate}

Training SNNs with gradient descent is complicated by the
non-differentiability of the spike generation
function~(Eq.~\ref{eq:spike_fn}): the Heaviside step has zero
derivative everywhere except at the threshold, blocking gradient
flow through the network. The surrogate gradient
method~\cite{neftci2019surrogate} resolves this by substituting
a smooth differentiable function $\hat{\sigma}'$ for the true
derivative during the backward pass:

\begin{equation}
    \frac{\partial S}{\partial U}
    \approx \hat{\sigma}'(U - U_{\mathrm{thr}}).
    \label{eq:surrogate}
\end{equation}

\noindent We use a tanh-based surrogate gradient implemented
in \texttt{snntorch}~\cite{eshraghian2023training}:

\begin{equation}
    \hat{\sigma}'(x) = 1 - \tanh^2(kx),
    \label{eq:surrogate_tanh}
\end{equation}

\noindent where $k$ controls the sharpness of the
approximation. Note that $\hat{\sigma}'(x)$ is the derivative
of $\tanh(kx)/k$, which closely approximates the Heaviside
step near the threshold and decays smoothly away from it.
Gradients are propagated through time via
spatio-temporal backpropagation~(STBP)~\cite{wu2018spatio},
which unrolls the network across $T$ timesteps and accumulates
gradients over the full simulation window.

\subsection{Output Decoding}
\label{subsec:decoding}

Output decoding defines the interface between binary spike
trains and the real-valued predictions required by downstream
tasks.

\paragraph{\textbf{Rate decoding}} uses the mean firing rate
$\bar{S} = \frac{1}{T}\sum_{t=1}^{T} S[t]$ as a continuous
output proxy. The class with the highest firing rate is taken
as the predicted category for classification tasks. Precision
is limited to $1/T$ for regression, making it unsuitable for
direct continuous value estimation.

\paragraph{\textbf{Membrane potential decoding}} removes the spike
threshold from the output layer, allowing the membrane potential
to accumulate continuously without resetting. The resulting
floating-point value is decoupled from $T$ and is well-suited
for regression. We adopt this for the vmem inference variant
of our model.

\paragraph{\textbf{Discretized regression}} converts a continuous
regression target into $C$ uniform bins, where $C$ is the
number of classes, transforming the problem into a
classification task~\cite{eshraghian2023training}.
The predicted class is obtained via:

\begin{equation}
    \hat{k} = \arg\max_{c}\;
    \mathrm{softmax}\!\left(
        \frac{1}{T}\sum_{t=1}^{T} S_{\mathrm{out},c}[t]
    \right),
    \label{eq:binned_cls}
\end{equation}

\noindent which is effective even at small $T$ since only a
relative ordering of firing rates is required. We apply this
for rotation prediction~(31 bins of $6^\circ$).

\paragraph{\textbf{Population-coded regression}} encodes a continuous
value through the collective mean firing rate of a local
$k \times k$ neuron cluster rather than the output of a single
neuron~\cite{pouget2000information}. This allows direct
spike-based regression without discretization and is the approach we adopt for bounding box dimension 
prediction, where the $\log_{10}$-transformed target 
dimensions $[\log_{10}(H), \log_{10}(W), \log_{10}(L)]$ 
are decoded from the mean firing rate of a local neuron 
population at each keypoint location, as described in 
Section~\ref{subsec:pop_box_loss}.

\section{Proposed Method}
\label{sec:method}

BEV-based CNN detectors achieve computational efficiency
through 2D convolutions on projected point cloud
representations, avoiding the cost of 3D volumetric
processing.
We aim to augment this computational efficiency with
inference energy efficiency by replacing dense activations
with binary spike trains.
A key departure from our prior CNN to SNN conversion
work~\cite{mohapatra2022spikili} is the adoption of direct
training with surrogate gradients. Conversion-based methods inherit
the accuracy ceiling of the source network and introduce
approximation errors that limit SNN performance. By
training end-to-end from scratch, the network is free to
learn spike-based representations optimised for the
detection task directly. Combined with careful network architecture design
and novel spike-domain loss functions, we aim to bring
SNN detection accuracy close to full-precision CNN baselines while maintaining substantial 
inference energy efficiency gains. The following subsections describe the
input representation, network architecture, and loss functions that together realise these goals.

\subsection{Input Preparation}
\label{subsec:input}
Raw LiDAR point clouds are inherently sparse and unstructured,
making direct neural network processing compute-intensive and
poorly suited to real-time embedded deployment.
Following~\cite{mohapatra2021bevdetnet, mohapatra2025lidar,
mohapatra2021limoseg, barrera2020birdnet+}, we project the 3D
point cloud into a 2D bird's eye view~(BEV) representation.
A $60\,\mathrm{m} \times 60\,\mathrm{m}$ ground-plane range is
discretized at a cell resolution of $\Delta\!=\!0.1875\,\mathrm{m}$,
yielding a spatial grid of $320 \times 320$ cells. For each 3D
point $\mathbf{p} = (x, y, z, i)$, the grid indices are:
\begin{equation}
    r = \left\lfloor\frac{x_{\max} - x}{\Delta}\right\rfloor,
    \qquad
    c = \left\lfloor\frac{y_{\max} - y}{\Delta}\right\rfloor,
    \label{eq:bev_proj}
\end{equation}
where $(x_{\max}, y_{\max})$ are the upper bounds of the covered
range. All $z$ values are shifted by $|z_{\min}|$ to start from
zero before feature computation.

Five geometric and reflectance features are computed per pillar
from all points mapped to it: maximum height~($z_{\max}$),
binary occupancy~($d$), mean reflectance intensity~($\bar{i}$),
minimum height~($z_{\min}$), and height standard
deviation~($\sigma_z$), computed via the Welford online
algorithm for numerical stability. These form the primary BEV
map $\mathbf{I}_{\mathrm{BEV}} \in \mathbb{R}^{320 \times 320
\times 5}$. All five channels are normalized to $[0,1]$ using
fixed physical bounds: heights by the shifted $z$ range, binary
occupancy is already in $\{0,1\}$, intensity is clipped to
$[0,1]$ directly from the Velodyne sensor range, and $\sigma_z$
is normalized by half the shifted $z$ range, since the standard
deviation of a bounded variable cannot exceed half its range.
Empty pillars are explicitly zeroed after normalization.

To preserve vertical structure within each pillar, the $z$-axis
is uniformly discretized into $K\!=\!6$ height bins spanning
ground level to $2.4\,\mathrm{m}$, covering the full height
range of passenger vehicles. An auxiliary binary occupancy map
$\mathbf{Z}_{\mathrm{BEV}} \in \{0,1\}^{320 \times 320 \times
6}$ records the presence of at least one point in each bin at
every pillar. The two representations are concatenated along the
channel dimension to form the full network input:
\begin{equation}
    \mathbf{X} =
    \left[\mathbf{I}_{\mathrm{BEV}} \;\|\;
          \mathbf{Z}_{\mathrm{BEV}}\right]
    \in \mathbb{R}^{320 \times 320 \times 11},
    \label{eq:input}
\end{equation}
giving $C\!=\!11$ input channels, consistent with the network
input shape reported in Table~\ref{tab:energy-detailed}.

\subsection{Network Architecture}
\label{subsec:arch}

\subsubsection{\textbf{Overview}}

The proposed network is an encoder-decoder architecture
following the U-Net~\cite{ronneberger2015u} design principle,
adapted for spiking computation. The input tensor $\mathbf{X}
\in \mathbb{R}^{320 \times 320 \times 11}$ is first processed
by a stem convolutional layer followed by a LIF neuron layer,
converting the continuous BEV map into a spike representation
via the self-coding strategy described in
Section~\ref{subsec:spike_coding}. All intermediate activations
throughout the network are binary spike trains $\mathbf{S} \in
\{0,1\}^{T \times B \times C \times H \times W}$ (where $B$ is
the batch size, $C$ the channel count, and $H \times W$ the
spatial resolution), processed over $T\!=\!13$ timesteps. 
The value of $T\!=\!13$ was determined empirically. Increasing
$T$ beyond 13 exceeded GPU memory capacity at a batch size
of~4 at $320\!\times\!320$ resolution. Additionally, longer
unrolled sequences increase the risk of vanishing gradients
during spatio-temporal backpropagation, as surrogate gradients
attenuate with each additional timestep. Training time also
scales linearly with $T$, making larger values impractical
within the available compute budget. $T\!=\!13$ represents
the largest value satisfying all three constraints
simultaneously.

\subsubsection{\textbf{Encoder}}

The encoder comprises four downsampling blocks~(DB1--DB4) that
progressively extract features at decreasing spatial
resolutions. Each DB contains two Conv$\,|\,$GN$\,|\,$LIF
stages followed by a channel-concatenation skip connection and
a strided pooling convolution. The first convolution uses a
$5\!\times\!5$ kernel to capture the spatial extent of objects
at each scale. The output of the second Conv$\,|\,$GN$\,|\,$LIF
stage is concatenated with the block input along the channel
dimension~($\otimes$), preserving the original feature map
before pooling. A final strided convolution halves the spatial
resolution to $H/2\!\times\!W/2$. Each DB therefore produces
two outputs: a full-resolution skip feature map passed to the
corresponding decoder block, and a half-resolution feature map
passed to the next encoder stage. Group normalisation~(GN) is
used throughout in place of batch normalisation~(BN), as it is
independent of batch size and behaves identically at training
and inference time---an important property for neuromorphic
deployment where batch size is typically one.

\subsubsection{\textbf{Decoder}}

The decoder comprises four upsampling blocks~(UB) arranged in a
bottom-up path from the bottleneck to full resolution. Each UB
receives the skip feature map from its corresponding encoder
level and the upsampled feature map from the previous decoder
stage, concatenates them~($\otimes$), and processes the joint
representation through a Conv$\,|\,$GN$\,|\,$LIF stage. Spatial
upsampling is performed by a transposed convolution followed by
GN, doubling the spatial resolution at each stage. This
hierarchical aggregation allows the decoder to combine
fine-grained spatial information from early encoder stages with
semantically rich features from the bottleneck, which is
particularly important for accurate keypoint localisation on
the sparse BEV map.

\subsubsection{\textbf{Output Heads}}

Three parallel spiking output heads operate on the
full-resolution decoder output. Each head consists of a single
Conv$\,|\,$LIF stage with no intermediate normalisation,
allowing the membrane potential to accumulate freely across
timesteps without normalisation interference.

\paragraph{Keypoint head} Predicts object-centre heatmaps from
which keypoint locations are decoded. Objects are represented
as Gaussian heatmap targets with unit peak at their projected
ground-plane centres, decaying spatially to suppress background
firing in the vicinity of object locations. The head is
supervised by the two-point temporal focal and masked Dice loss
described in Section~\ref{subsubsec:kp_loss}.

\paragraph{Box head} Predicts bounding box dimensions
$[\log_{10}(h),\,\log_{10}(w),\,\log_{10}(l)]$ at keypoint
locations. The $\log_{10}$ transformation maps typical car
dimensions~(0.5--10\,m) to $[0,1]$, making the regression
target compatible with spike-rate outputs. Dimensions are
supervised by the population-coded spike regression loss
described in Section~\ref{subsec:pop_box_loss}.

\paragraph{Rotation head} Predicts heading orientation as a
classification over 31 uniform bins of $6^{\circ}$ each,
covering $[0,\pi]$ to exploit the rotational symmetry of the
BEV representation. A bin width of $6^{\circ}$ provides
sufficient angular resolution for vehicle orientation
estimation in BEV while remaining reliably learnable within
$T\!=\!13$ timesteps. Label smoothing of $\varsigma\!=\!0.1$
is applied to reflect the geometric similarity of adjacent
orientation bins. The head is supervised by the masked
cross-entropy loss described in
Section~\ref{subsubsec:rot_loss}.

\begin{figure*}[!t]
    \centering
    \includegraphics[width=0.8\textwidth]{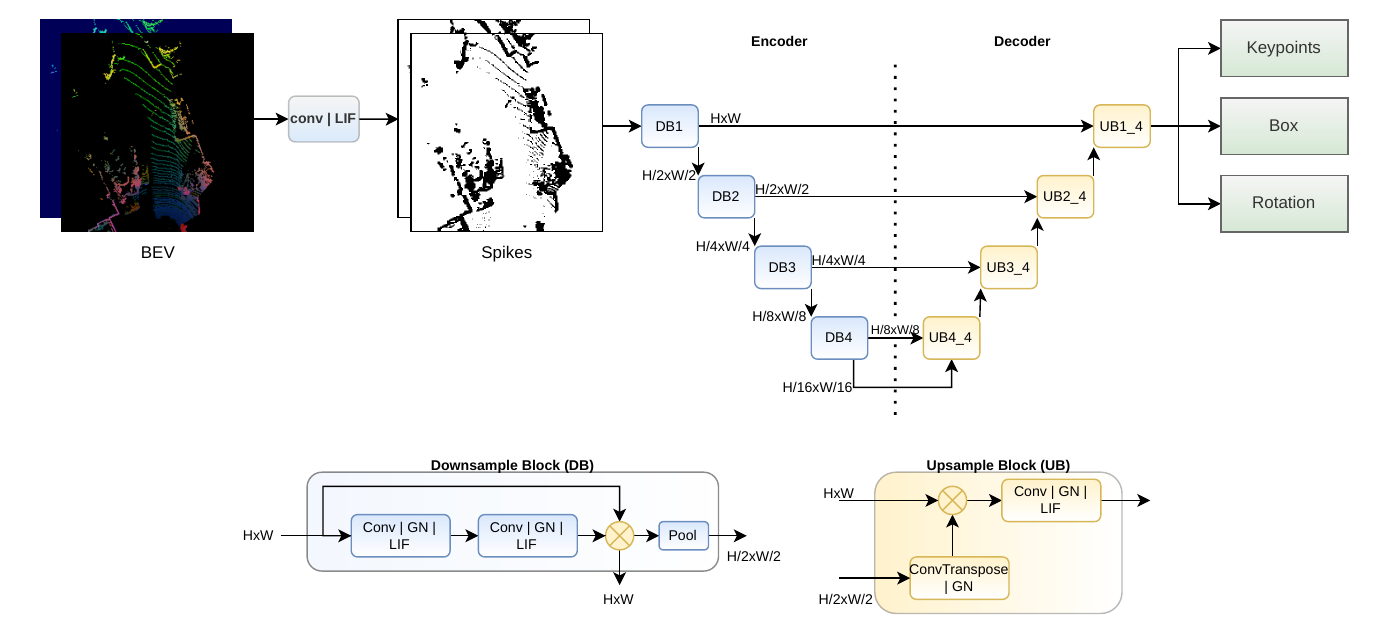}
    \caption{\textbf{Network architecture (top)}: Downsampling Blocks~(DB)
        extract features and reduce spatial resolution while
        Upsampling Blocks~(UB) aggregate multi-scale features and
        restore spatial resolution. The convolutional layers act as
        synaptic connections and feature extractors between the
        Leaky Integrate-and-Fire~(LIF) neuron layers.}
    \label{fig:network_arch}
\end{figure*}

\subsection{Losses}

\subsubsection{\textbf{Keypoint Detection Loss}}
\label{subsubsec:kp_loss}

The keypoint detection head produces spike trains
$\mathbf{S}_{\mathrm{kp}} \in \{0,1\}^{T \times B \times 1
\times H \times W}$. Object centres are represented as Gaussian
heatmap targets $\mathbf{Y}^{*}_{\mathrm{kp}} \in [0,1]^{B
\times H \times W}$, with unit peak at each object centre
decaying spatially with a Gaussian kernel. Keypoint peaks
typically constitute less than $1\%$ of the spatial map,
creating severe class imbalance between object-centre and
background pixels. The proposed loss $\mathcal{L}_{\mathrm{kp}}$
addresses this through two components, both operating on
temporally aggregated spike rates. Table~\ref{tab:kp_notation}
defines the notation used throughout this section.

\begin{table}[!t]
\renewcommand{\arraystretch}{1.2}
\caption{\textbf{Notation Summary for $\mathcal{L}_{\mathrm{kp}}$}}
\label{tab:kp_notation}
\centering
\begin{tabular}{ll}
\toprule
\textbf{Symbol} & \textbf{Description} \\
\midrule
$T$                              & Number of timesteps \\
$\alpha_t$                       & Early window fraction \\
$T_e$                            & Early window steps
                                   $\lfloor\alpha_t T\rfloor$ \\
$k$                              & ROI dilation radius \\
$\mathbf{S}_{\mathrm{kp}}^{(t)}$ & Binary spike tensor at timestep $t$ \\
$\hat{\mathbf{R}}_e$             & Early-window mean firing rate map \\
$\hat{\mathbf{R}}_f$             & Full-window mean firing rate map \\
$\mathbf{Y}^{*}_{\mathrm{kp}}$   & Ground-truth Gaussian heatmap \\
$\alpha,\,\beta$                 & Focal loss exponents \\
$w_e,\, w_f$                     & Early and full window focal weights \\
$\mathbf{M}_{\mathrm{roi}}$      & Dilated keypoint ROI mask \\
$\tilde{\mathbf{r}}_b,\,
 \tilde{\mathbf{y}}^{*}_{\mathrm{kp},b}$ & Masked rate and GT for item $b$ \\
$\epsilon$                       & Dice smoothing constant \\
$\lambda_{\mathrm{d}}$           & Masked Dice loss weight \\
$e,\; e_{\mathrm{gate}}$         & Current epoch; Dice activation epoch \\
$\mathcal{L}_{\mathrm{kp}}$      & Keypoint detection loss \\
\bottomrule
\end{tabular}
\end{table}

\paragraph{Temporal Rate Aggregation.}
Applying a loss directly to per-timestep spikes
$\mathbf{S}^{(t)} \in \{0,1\}$ produces unstable gradients.
Instead, spikes are aggregated over two temporal windows to
obtain smooth rate estimates in $[0,1]$. Let
$T_e = \lfloor \alpha_t T \rfloor$ denote the early window
boundary. The early and full window spike rates are:

\begin{align}
    \hat{\mathbf{R}}_e
        &= \frac{1}{T_e}
           \sum_{t=1}^{T_e} \mathbf{S}_{\mathrm{kp}}^{(t)}
        \;\in\; [0,1]^{B \times 1 \times H \times W},
        \label{eq:rate_early} \\
    \hat{\mathbf{R}}_f
        &= \frac{1}{T}
           \sum_{t=1}^{T} \mathbf{S}_{\mathrm{kp}}^{(t)}
        \;\in\; [0,1]^{B \times 1 \times H \times W}.
        \label{eq:rate_full}
\end{align}

\noindent $\hat{\mathbf{R}}_f$ provides the most reliable
estimate over the full temporal window, while
$\hat{\mathbf{R}}_e$ ensures early timesteps receive direct
gradient supervision rather than only diluted gradients through
the full temporal mean.

\paragraph{Two-Point Gaussian Focal Loss.}
Standard weighted BCE requires an explicit positive weight estimated from training-set statistics. 
We instead adopt the focal loss~\cite{lin2017focal} formulation of CornerNet~\cite{law2018cornernet}, adapted for Gaussian heatmap targets, which handles class imbalance implicitly without requiring explicit positive weights.
For a predicted rate map $\hat{p}$ and Gaussian target $y \in [0,1]$, the per-pixel focal loss is:

\begin{equation}
    \ell(y, \hat{p}) =
    \begin{cases}
        -(1 - \hat{p})^{\alpha} \log \hat{p}
            & \text{if } y = 1, \\[4pt]
        -(1 - y)^{\beta}\, \hat{p}^{\alpha} \log(1 - \hat{p})
            & \text{otherwise,}
    \end{cases}
    \label{eq:focal_pixel}
\end{equation}

\noindent where $\alpha\!=\!2$ focuses gradient on hard
positives, and $\beta\!=\!4$ down-weights pixels near object
centres that are partially positive due to Gaussian
spread---preventing them from being penalised as true
background. The total is normalised by the number of positive
pixels, clamped to a minimum of 1 to handle empty scenes.
The two-point focal loss is:

\begin{equation}
    \mathcal{L}_{\mathrm{FL}}
    = \frac{w_e\,\mathcal{L}_{\mathrm{FL}}(\hat{\mathbf{R}}_e,\,
                              \mathbf{Y}^{*}_{\mathrm{kp}})
           + w_f\,\mathcal{L}_{\mathrm{FL}}(\hat{\mathbf{R}}_f,\,
                              \mathbf{Y}^{*}_{\mathrm{kp}})}
           {w_e + w_f},
    \label{eq:focal_twopoint}
\end{equation}

\noindent where $w_e < w_f$ reflect that the early window is
noisier and should contribute less to the primary accuracy
signal. Normalisation by $(w_e + w_f)$ maintains a stable
absolute loss scale.

\paragraph{Masked Soft Dice Loss.}
The full-window rate map $\hat{\mathbf{R}}_f$ is additionally
supervised by a masked soft Dice loss~\cite{milletari2016vnet}.
Standard Dice on a sparse BEV map is dominated by background
firing in the denominator, rendering it insensitive to whether
keypoint peaks are correctly predicted. The Dice computation is
therefore restricted to a region of interest
$\mathbf{M}_{\mathrm{roi}}$ obtained by dilating the binary
peak mask $(\mathbf{Y}^{*}_{\mathrm{kp}}\!=\!1)$ via max
pooling:

\begin{equation}
    \mathbf{M}_{\mathrm{roi}}
    = \mathrm{MaxPool}\!\left(
        \mathbb{1}[\mathbf{Y}^{*}_{\mathrm{kp}}\!=\!1];\;
        \tilde{k},\; \lfloor \tilde{k}/2 \rfloor
      \right),
    \label{eq:roi}
\end{equation}

\noindent where $\tilde{k}$ is the smallest odd integer $\geq k$,
ensuring valid same-size padding. The masked prediction and
target are:

\begin{align}
    \tilde{\mathbf{r}}_b
        &= \hat{\mathbf{r}}_b \odot \mathbf{m}_{\mathrm{roi},b},
        \label{eq:masked_r} \\
    \tilde{\mathbf{y}}^{*}_{\mathrm{kp},b}
        &= \mathbf{y}^{*}_{\mathrm{kp},b} \odot \mathbf{m}_{\mathrm{roi},b},
        \label{eq:masked_y}
\end{align}

\noindent where $\odot$ denotes element-wise multiplication and
lower-case variables denote spatially-flattened vectors of their
corresponding maps. The masked Dice loss is:

\begin{equation}
    \mathcal{L}_{\mathrm{Dice}}
    = 1 - \frac{1}{B}\sum_{b=1}^{B}
      \frac{2\langle\tilde{\mathbf{r}}_{b},\,
                    \tilde{\mathbf{y}}^{*}_{\mathrm{kp},b}\rangle + \epsilon}
           {\|\tilde{\mathbf{r}}_{b}\|_1
           + \|\tilde{\mathbf{y}}^{*}_{\mathrm{kp},b}\|_1 + \epsilon},
    \label{eq:masked_dice}
\end{equation}

\noindent where $\epsilon > 0$ stabilizes gradients when the
ROI is near-empty~\cite{sudre2017generalised}.

\paragraph{Combined Loss with Epoch Gating.}
In early training the network has not yet learned to fire at
keypoint locations, so the masked Dice numerator is near zero
and its gradient is uninformative. We therefore gate
$\mathcal{L}_{\mathrm{Dice}}$ by training epoch $e$:

\begin{equation}
    \mathcal{L}_{\mathrm{kp}} =
    \begin{cases}
        \mathcal{L}_{\mathrm{FL}}
            & e < e_{\mathrm{gate}}, \\[4pt]
        \mathcal{L}_{\mathrm{FL}}
        + \lambda_{\mathrm{d}}\,\mathcal{L}_{\mathrm{Dice}}
            & e \geq e_{\mathrm{gate}},
    \end{cases}
    \label{eq:kp_total}
\end{equation}

\noindent where $\lambda_{\mathrm{d}} \ll 1$ ensures the
masked Dice term provides only structural spatial guidance
without destabilising the primary focal signal. In our
experiments, $e_{\mathrm{gate}}$ is set to epoch~40 out of
200 total training epochs, corresponding to the first $20\%$
of training, by which point the focal loss has established a
meaningful firing pattern at object centres.

\subsubsection{\textbf{Population-Coded Box Regression Loss}}
\label{subsec:pop_box_loss}

Estimating continuous physical dimensions---namely length $l$, width $w$, and
height $h$---from the sparse binary spike trains produced by a spiking neural
network (SNN) presents a fundamental challenge: individual neurons emit
stochastic, all-or-nothing pulses that are poorly suited to representing
real-valued regression targets in isolation. To address this, we propose a
\textit{Population-Coded Box Regression Loss} ($\mathcal{L}_{\mathrm{box}}$)
grounded in the principle of population coding~\cite{pouget2000}, wherein a local
$k \times k$ cluster of neurons collectively encodes a single continuous
quantity through their aggregate firing activity.

\paragraph{Temporal Rate Integration}

Let $\mathbf{S}^{(t)} \in \{0,1\}^{B \times 3 \times H \times W}$ denote the
binary spike tensor emitted by the box regression head at timestep $t$, where
$B$ is the batch size, $3$ corresponds to the three regression targets
$\{l, w, h\}$, and $H \times W$ is the spatial resolution of the Bird's-Eye
View~(BEV) feature map. Over a time window of $T$ timesteps, we compute
the \textit{mean firing rate map}:

\begin{equation}
    \hat{\mathbf{R}}_{\mathrm{box}} = \frac{1}{T} \sum_{t=1}^{T} \mathbf{S}^{(t)}
    \;\in\; [0,\,1]^{B \times 3 \times H \times W},
    \label{eq:rate_integration}
\end{equation}

\noindent which converts the discrete spike train into a continuous
intensity representation amenable to regression supervision.

\paragraph{Spatial Population Integration}

Rather than supervising each spatial neuron independently, we compute the
\textit{population consensus} within a local $k \times k$ neighborhood via
average pooling. Let $b \in \{1,\ldots,B\}$ denote the batch index,
$c \in \{l,w,h\}$ the regression channel, and $(i,j) \in \{1,\ldots,H\}
\times \{1,\ldots,W\}$ the spatial pixel location. The population-pooled
prediction is then:

\begin{equation}
    \hat{\mathbf{R}}_{\mathrm{pop}}[b,c,i,j]
    = \frac{1}{k^2}
      \sum_{(\delta_i,\,\delta_j)\,\in\,\mathcal{N}_k}
      \hat{\mathbf{R}}_{\mathrm{box}}\!\left[b,\,c,\,i+\delta_i,\,j+\delta_j\right],
    \label{eq:pop_integration}
\end{equation}

\noindent where $\mathcal{N}_k = \{-\lfloor k/2 \rfloor, \ldots,
\lfloor k/2 \rfloor\}^2$ is the $k \times k$ local neighborhood centered
at pixel $(i,j)$. This pooling operation encodes the key insight of
population coding: the regression target is satisfied by the \textit{collective
mean} of the cluster, permitting individual neurons to exhibit firing variance
or intermittent silence while the population as a whole converges to the
desired value. Critically, this relaxation stabilizes gradient flow during
surrogate-gradient backpropagation ~\cite{neftci2019}, where individual spike
gradients are noisy approximations of the true gradient.

\paragraph{Masked Regression Objective}

Since box dimensions are only defined at object center locations, the
regression loss is restricted to pixels identified as keypoints by the
ground-truth keypoint map $\mathbf{Y}^{*}_{\mathrm{kp}} \in \{0,1\}^{B \times H
\times W}$. Let the spatial mask be:

\begin{equation}
    \mathbf{M}[b,c,i,j] =
    \begin{cases}
        1 & \text{if } \mathbf{Y}^{*}_{\mathrm{kp}}[b,i,j] > 0 \\
        0 & \text{otherwise,}
    \end{cases}
    \label{eq:mask}
\end{equation}

\noindent expanded across the channel dimension $c \in \{l,w,h\}$. The final
loss is the $\ell_1$ norm computed exclusively over the masked keypoint
population responses:

\begin{equation}
    \mathcal{L}_{\mathrm{box}}
    = \sum_{b,c,i,j}
      \mathbf{M}[b,c,i,j]
      \cdot
      \left|
          \hat{\mathbf{R}}_{\mathrm{pop}}[b,c,i,j]
          \,-\,
          \mathbf{Y}^{*}_{\mathrm{box}}[b,c,i,j]
      \right|,
    \label{eq:box_loss}
\end{equation}

\noindent where $\mathbf{Y}^{*}_{\mathrm{box}} \in \mathbb{R}^{B \times 3 \times H \times W}$
is the ground-truth box dimension map. The $\ell_1$ norm is chosen over the
mean-squared error for its robustness to outliers in physical dimension
estimates and its well-behaved gradient magnitude near zero, both of which
are critical properties for stable convergence in surrogate-gradient
training~\cite{lee2016}.

If a batch contains no keypoint pixels---corresponding to an empty
scene---the loss degenerates gracefully to:

\begin{equation}
    \mathcal{L}_{\mathrm{box}} = \mathbf{0},
    \label{eq:empty_loss}
\end{equation}

\noindent while preserving the autograd computation graph to prevent
silent gradient failures.

\begin{figure}[!t]
\centering
\includegraphics[width=\columnwidth]{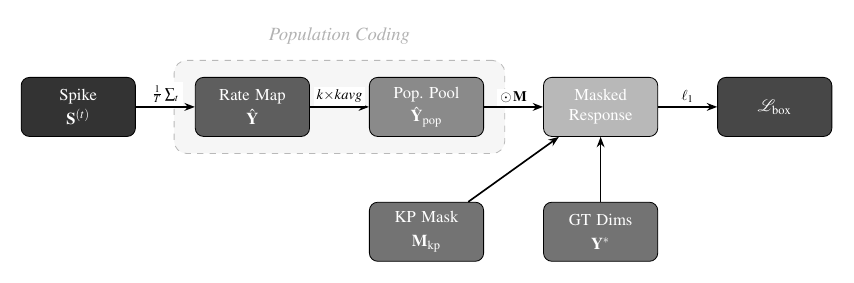}
\caption{
    \textbf{Pipeline of the proposed Population-Coded Box Regression Loss
    $\mathcal{L}_{\mathrm{box}}$}:
    Binary spike trains $\mathbf{S}^{(t)}$ are temporally integrated into a
    continuous rate map $\hat{\mathbf{R}}_{\mathrm{box}}$ (Eq.~\ref{eq:rate_integration}),
    then spatially pooled over a $k{\times}k$ neighborhood to form the
    population consensus $\hat{\mathbf{R}}_{\mathrm{pop}}$
    (Eq.~\ref{eq:pop_integration}).
    The $\ell_1$ regression loss (Eq.~\ref{eq:box_loss}) is computed solely
    at object centers defined by the keypoint mask $\mathbf{Y}^{*}_{\mathrm{kp}}$.
}
\label{fig:pop_box_loss}
\end{figure}

\begin{table}[!t]
\renewcommand{\arraystretch}{1.25}
\caption{Notation Summary for $\mathcal{L}_{\mathrm{box}}$}
\label{tab:notation}
\centering
\begin{tabular}{lp{5.5cm}}
\toprule
\textbf{Symbol} & \textbf{Description} \\
\midrule
$T$                              & Number of timesteps \\
$k$                              & Spatial extent of the neuron population cluster \\
$\mathbf{S}^{(t)}$               & Binary spike tensor at timestep $t$ \\
$\hat{\mathbf{R}}_{\mathrm{box}}$& Temporal mean firing rate map \\
$\hat{\mathbf{R}}_{\mathrm{pop}}$& Population-pooled prediction map \\
$\mathbf{Y}^{*}_{\mathrm{box}}$  & Ground-truth box dimension map $\{l, w, h\}$ \\
$\mathbf{Y}^{*}_{\mathrm{kp}}$   & Binary keypoint mask \\
$\mathbf{M}$                     & Channel-expanded keypoint mask \\
$\mathcal{L}_{\mathrm{box}}$     & Population-coded box regression loss \\
\bottomrule
\end{tabular}
\end{table}

\subsubsection{\textbf{Rotation Classification Loss}}
\label{subsubsec:rot_loss}

Rotation angles are mapped to $[0, \pi]$ and uniformly
discretized into $C\!=\!31$ bins of $6^\circ$ each, converting
heading estimation into a classification task. The rotation head
produces spike trains
$\mathbf{S}_{\mathrm{rot}} \in \{0,1\}^{T \times B \times C
\times H \times W}$, with one channel per orientation bin across
the full BEV spatial map. Since rotation is only defined at
object center locations, the loss is restricted to keypoint
pixels identified by the ground-truth mask
$\mathbf{Y}^{*}_{\mathrm{kp}} \in \{0,1\}^{B \times H \times W}$.

The temporal mean firing rate across all $T$ timesteps forms
the class logits:

\begin{equation}
    \hat{\mathbf{R}}_{\mathrm{rot}}
    = \frac{1}{T} \sum_{t=1}^{T}
      \mathbf{S}_{\mathrm{rot}}^{(t)}
    \;\in\; [0,1]^{B \times C \times H \times W}.
    \label{eq:rot_rate}
\end{equation}

\noindent Raw spike rates do not form a valid categorical
distribution across bins. A softmax normalization is therefore
applied across the channel dimension before computing the
cross-entropy, which is handled internally by the cross entropy (CE) loss
in the implementation. Let
$\mathbf{y}^{*}_{\mathrm{rot}} \in \{0,\ldots,C-1\}^{B \times
H \times W}$ denote the ground-truth bin index map, and let
$\mathbf{w}_c \in \mathbb{R}^C$ be a uniform per-class weight
vector that upweights rotation bins relative to background to
compensate for the spatial sparsity of object centers. Restricting
to keypoint locations via Boolean indexing, the rotation loss is:

\begin{equation}
    \mathcal{L}_{\mathrm{rot}}
    = \mathrm{CE}\!\left(
        \hat{\mathbf{R}}_{\mathrm{rot}}\big|_{\mathrm{kp}},\;
        \mathbf{y}^{*}_{\mathrm{rot}}\big|_{\mathrm{kp}},\;
        \mathbf{w}_c
      \right),
    \label{eq:rot_loss}
\end{equation}

\noindent where $(\cdot)|_{\mathrm{kp}}$ denotes extraction of
predictions and targets at pixels where
$\mathbf{Y}^{*}_{\mathrm{kp}}\!=\!1$, and label smoothing
$\varsigma\!=\!0.1$ is applied to reflect the geometric
similarity of adjacent orientation bins~\cite{szegedy2016rethinking}.


\begin{table}[!htbp]
\centering
\caption{\textbf{Car Detection Average Precision (AP) on KITTI Validation Set}}
\label{tab:my-table-AP}
\setlength{\tabcolsep}{4pt}
\footnotesize
\begin{tabular}{c|l|rrr|rrr}
\toprule
\multirow{2}{*}{Type} & \multirow{2}{*}{Model} & \multicolumn{3}{c|}{AP @ IoU=0.5} & \multicolumn{3}{c}{AP @ IoU=0.7} \\
 & & \cellcolor[HTML]{9AFF99}Easy & \cellcolor[HTML]{FFFFC7}Mod. & \cellcolor[HTML]{FFCCC9}Hard & \cellcolor[HTML]{9AFF99}Easy & \cellcolor[HTML]{FFFFC7}Mod. & \cellcolor[HTML]{FFCCC9}Hard \\
\midrule
\multirow{4}{*}{\rotatebox[origin=c]{90}{\textbf{CNN}}}
  & PointPillars~\cite{lang2019pointpillars} & 90.78 & 90.18 & 89.46 & 88.32 & 86.10 & 79.83 \\
  & PointRCNN~\cite{shi2019pointrcnn}        & 90.20 & 89.63 & 89.39 & 92.13 & 87.39 & 82.72 \\
  & Pixor~\cite{yang2018pixor}               & 89.62 & 83.45 & 80.11 & 86.79 & 80.75 & 76.60 \\
  & BEVDetNet~\cite{mohapatra2021bevdetnet}  & 87.82 & 87.78 & 87.25 & 82.46 & 77.90 & 77.45 \\
\midrule
\multirow{5}{*}{\rotatebox[origin=c]{90}{\textbf{SNN}}}
  & DeepSCNN~\cite{lian2020deep}                    &     --- &     --- &     --- & 71.76 & 67.43 & 65.63 \\
  & SpikeCloudNet~\cite{neumeier2025spikeclouds}    &     --- &     --- &     --- & 89.80 & 84.10 & 82.70 \\
  & SpikiLi~\cite{mohapatra2022spikili}             & 86.37 & 78.65 & 78.76 & 82.71 & 75.49 & 68.68 \\
  & \textbf{vmem (ours)}  & \textbf{92.05} & \textbf{87.04} & \textbf{86.51} & \textbf{90.11} & \textbf{85.50} & \textbf{82.80} \\
  & \textbf{spike (ours)} &          88.51 &          85.72 &          83.14 &          85.18 &          80.20 &          79.14 \\
\bottomrule
\end{tabular}
\end{table}

\begin{table}[!htbp]
\centering
\caption{\textbf{Impact of Input Spike Coding on Average Precision}}
\label{tab:input-coding}
\setlength{\tabcolsep}{10pt}
\footnotesize
\begin{tabular}{l|ccc}
\toprule
\multirow{2}{*}{\textbf{Encoding}} & \multicolumn{3}{c}{\textbf{AP @ IoU=0.5}} \\ \cmidrule(l){2-4}
 & \cellcolor[HTML]{9AFF99}Easy & \cellcolor[HTML]{FFFFC7}Mod. & \cellcolor[HTML]{FFCCC9}Hard \\
\midrule
Poisson / rate   & 87.14 & 85.83 & 82.21 \\
Temporal z-axis  & 78.05 & 75.44 & 70.98 \\
Latency          & 34.62 & 21.30 & 10.87 \\
Self-coding      & \textbf{92.05} & \textbf{87.04} & \textbf{86.51} \\
\bottomrule
\end{tabular}
\end{table}

\begin{table*}[tb]
\centering
\caption{\textbf{Block-wise synaptic operation energy comparison}: SNN vs.\ equivalent CNN. Energy estimated using 45\,nm technology constants ($E_\text{AC} = 0.9$\,pJ, $E_\text{MAC} = 4.6$\,pJ) at $T = 13$ timesteps and input resolution $320 \times 320$. FR = mean firing rate; Sp. = sparsity ($1 - \text{FR}$); CNN/SNN = energy ratio (higher is better for SNN).}
\label{tab:energy-detailed}
\setlength{\tabcolsep}{5pt}
\renewcommand{\arraystretch}{1.2}
\footnotesize
\begin{tabular}{clcrrrrrrrrr}
\toprule
\multirow{2}{*}{\textbf{Stage}} & \multirow{2}{*}{\textbf{Block}} & \multirow{2}{*}{\textbf{Input (C$\times$H$\times$W)}} & \multirow{2}{*}{\textbf{MACs}} & \multirow{2}{*}{\textbf{FR}} & \multirow{2}{*}{\textbf{Sp.}} & \multicolumn{2}{c}{\textbf{Energy ($\mu$J)}} & \multicolumn{2}{c}{\textbf{Energy Share (\%)}} & \multirow{2}{*}{\textbf{CNN/SNN}} \\
\cmidrule(lr){7-8} \cmidrule(lr){9-10}
 & & & & & & \textbf{SNN} & \textbf{CNN} & \textbf{SNN} & \textbf{CNN} & \\
\midrule
\multirow{2}{*}{\rotatebox[origin=c]{90}{\textbf{Input}}}
  & Stem      & $11\times320\times320$  &       162,201,600 & 0.0183 & 0.9817 &     34.7140 &    746.1274 &  0.04 &  0.24 & 21.49 \\
\cmidrule(l){2-11}
\multirow{5}{*}{\rotatebox[origin=c]{90}{\textbf{Encoder}}}
  & DB1       & $16\times320\times320$  &  2,785,280,000 & 0.0744 & 0.9256 &  2425.4326 &  12812.2880 &  2.55 &  4.04 &  5.28 \\
  & DB2       & $48\times160\times160$  &  3,632,332,800 & 0.0873 & 0.9127 &  3710.9472 &  16708.7309 &  3.90 &  5.27 &  4.50 \\
  & DB3       & $112\times80\times80$   &  4,066,918,400 & 0.0682 & 0.9318 &  3244.3032 &  18707.8246 &  3.41 &  5.90 &  5.77 \\
  & DB4       & $240\times40\times40$   &  4,286,976,000 & 0.0907 & 0.9093 &  4550.8497 &  19720.0896 &  4.78 &  6.22 &  4.33 \\
\cmidrule(l){2-11}
  & \textit{Subtotal} & ---            & 14,771,507,200 & 0.0806 & 0.9194 & 13931.5329 &  67948.9331 & 14.64 & 21.44 &  4.88 \\
\cmidrule(l){2-11}
\multirow{5}{*}{\rotatebox[origin=c]{90}{\textbf{Decoder}}}
  & UB4\_4    & $496\times20\times20$   &  8,659,763,200 & 0.1231 & 0.8769 & 12475.0975 &  39834.9107 & 13.11 & 12.57 &  3.19 \\
  & UB3\_4    & $496\times40\times40$   & 16,472,473,600 & 0.1414 & 0.8586 & 27253.3480 &  75773.3786 & 28.64 & 23.91 &  2.78 \\
  & UB2\_3    & $240\times80\times80$   & 14,981,529,600 & 0.0933 & 0.9067 & 16346.0321 &  68915.0362 & 17.18 & 21.74 &  4.22 \\
  & UB1\_2    & $112\times160\times160$ & 12,215,910,400 & 0.1295 & 0.8705 & 18507.2248 &  56193.1878 & 19.45 & 17.73 &  3.04 \\
\cmidrule(l){2-11}
  & \textit{Subtotal} & ---            & 52,329,676,800 & 0.1218 & 0.8782 & 74581.7025 & 240716.5133 & 78.37 & 75.95 &  3.23 \\
\cmidrule(l){2-11}
\multirow{4}{*}{\rotatebox[origin=c]{90}{\textbf{Output}}}
  & Keypoint  & $48\times320\times320$  &    532,070,400 & 0.3485 & 0.6515 &  2169.7737 &   2447.5238 &  2.28 &  0.77 &  1.13 \\
  & Box       & $48\times320\times320$  &    534,528,000 & 0.3465 & 0.6535 &  2166.8515 &   2458.8288 &  2.28 &  0.78 &  1.13 \\
  & Rotation  & $48\times320\times320$  &    568,934,400 & 0.3434 & 0.6566 &  2285.8559 &   2617.0982 &  2.40 &  0.83 &  1.14 \\
\cmidrule(l){2-11}
  & \textit{Subtotal} & ---            &  1,635,532,800 & 0.3461 & 0.6539 &  6622.4811 &   7523.4509 &  6.96 &  2.37 &  1.14 \\
\midrule
\rowcolor[HTML]{DAE8FC}
\multicolumn{1}{c}{} & \textbf{Total} & --- & 68,898,918,400 & 0.1181 & 0.8819 & 95170.4304 & 316935.0246 & 100.00 & 100.00 & \textbf{3.33} \\
\bottomrule
\end{tabular}
\end{table*}

\section{Experiments and Results}
\label{sec:experiments}

\subsection{Dataset, Evaluation Protocol, and Training}

\begin{figure*}[!t]
    \centering
    \includegraphics[width=\textwidth]{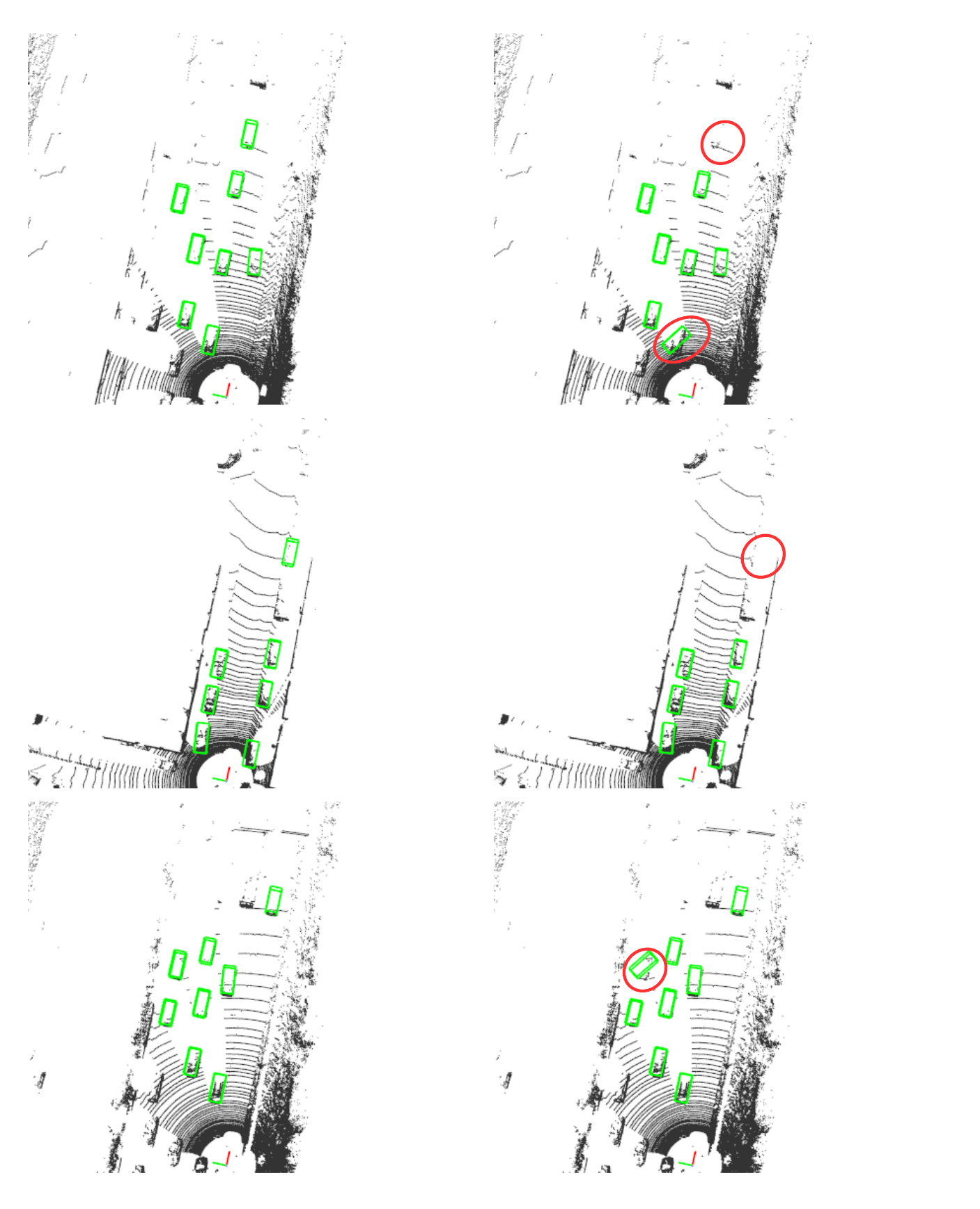}
    \caption{\textbf{Qualitative comparison of 3D object detection on the
        KITTI validation set across three diverse scenes}: Green boxes
        denote network predictions. The left column demonstrates the
        continuous \textit{vmem} readout, while the right column
        demonstrates the fully binary \textit{spike} readout. Red
        circles highlight characteristic failure modes induced by
        purely binary spike inference, including false negatives on
        distant, sparse objects (top and middle rows) and orientation
        misalignment (bottom row).}
    \label{fig:qualitative_comp}
\end{figure*}

We evaluate on the KITTI object detection benchmark~\cite{geiger2013vision}
using the official train/validation split~\cite{chen2015multi}. Horizontal
flipping augmentation is applied selectively to training frames, yielding a
total of 6,000 training samples. All models are trained for 200 epochs using
Adam~\cite{kingma2014adam} with cosine annealing with warm
restarts~\cite{loshchilov2016sgdr} and gradient clipping at
$\|\nabla\|_{\max}\!=\!1.0$ to stabilize surrogate-gradient
training~\cite{neftci2019}. SNNs are implemented in
\texttt{snntorch}~\cite{eshraghian2023training}.

We report AP for the \textit{Car} class at two IoU thresholds across
\textit{Easy}, \textit{Moderate}, and \textit{Hard} difficulty levels.
IoU$\,{=}\,0.5$ is the operationally relevant threshold for ADAS
applications---including collision avoidance, emergency braking, and path
planning---where coarse but reliable localization is sufficient for safe
decision-making~\cite{caesar2020nuscenes}. IoU$\,{=}\,0.7$ demands tighter
localization and is included for direct comparison with prior work.

\subsection{Temporal Encoding Strategy}
A key design choice in SNN-based perception is how continuous sensor
data is converted into spike trains. We evaluate four established
strategies~\cite{auge2021survey} in Table~\ref{tab:input-coding}:
Poisson/rate coding, temporal z-axis encoding, latency coding, and
self-coding~\cite{rueckauer2017conversion}. In self-coding, the same
BEV frame is presented identically over $T\!=\!13$ timesteps, allowing
LIF neurons to integrate membrane potential across repeated
presentations and converge to a stable internal representation. While
this involves frame repetition in offline evaluation, in a deployed
system receiving a continuous LiDAR stream each timestep naturally
corresponds to a distinct sensor frame, providing a genuine temporal
axis without duplication. The static-input results therefore represent
a conservative lower bound on real-world performance.

As shown in Table~\ref{tab:input-coding}, self-coding achieves the
highest performance at $92.05$/$87.04$/$86.51$ AP at $IoU\!=\!0.5$.
Poisson/rate coding is the closest alternative at
$87.14$/$85.83$/$82.21$, trailing by approximately $1$--$4$ AP
points---a modest but consistent gap attributable to stochastic
firing variance introduced by Poisson sampling, which adds noise to
the spatial feature representation. Temporal z-axis encoding trails
further at $78.05$/$75.44$/$70.98$, reflecting the loss of
within-cell height distribution inherent in collapsing the vertical
LiDAR axis into a temporal sequence.

Latency coding collapses to near-chance performance
($34.62$/$21.30$/$10.87$). This is not a tuning failure but a
structural incompatibility: latency coding permits at most one spike
per neuron over the entire temporal window, inducing extreme
activation sparsity from the first layer onward. On dense BEV inputs
where most spatial cells carry valid non-zero features, this
one-spike constraint destroys the spatial density the architecture
depends on and renders surrogate gradient training ineffective, as
near-zero gradients propagate through the large fraction of neurons
that never reach threshold. Self-coding avoids this pathology by
allowing unrestricted firing across timesteps, enabling reliable
gradient flow throughout the full spatial extent of the encoder-decoder.

\subsection{Membrane Potential and Spike Inference Variants}
\label{subsec:variants}
We evaluate two inference variants that together bound the
accuracy--deployability trade-off of the proposed architecture.
Both variants share identical weights; they differ only in how
the output heads produce predictions.

\subsubsection{\textbf{Membrane Potential Variant (vmem)}}
The vmem variant reads the final membrane potential $V_{\mathrm{mem}}$
of the last LIF layer, a continuous quantity accumulating evidence
over all $T$ timesteps. All intermediate layers remain fully spiking,
preserving the MAC-to-AC reduction throughout the feature
extractor~\cite{roy2019towards}. vmem is compatible with neuromorphic
platforms that expose internal membrane state, such as Intel
Loihi~2~\cite{orchard2021loihi} and IBM
TrueNorth~\cite{merolla2014million} in monitoring mode.

vmem~(ours) achieves $92.05$/$87.04$/$86.51$ AP at $IoU\!=\!0.5$
and $90.11$/$85.50$/$82.80$ at $IoU\!=\!0.7$
(Easy/Moderate/Hard, Table~\ref{tab:my-table-AP}), establishing a
new state-of-the-art among SNN-based detectors across all difficulty
tiers at $IoU\!=\!0.7$ and exceeding all CNN baselines on the Easy
metric at $IoU\!=\!0.5$. This demonstrates that a spiking backbone
with a continuous readout can match and in some cases surpass
full-precision CNN counterparts.

\subsubsection{\textbf{Fully Spiking Variant (spike)}}
The spike variant replaces the $V_{\mathrm{mem}}$ readout with
predictions derived from the temporal mean firing rate over $T$
timesteps. Every layer operates exclusively on binary spike trains
$\mathbf{S} \in \{0,1\}^{T \times H \times W}$, making spike
directly deployable on any event-driven neuromorphic processor
without modification, including platforms that do not expose
membrane state such as SpiNNaker~\cite{furber2014spinnaker} and
BrainScaleS~\cite{schemmel2010wafer}.

spike~(ours) achieves $88.51$/$85.72$/$83.14$ AP at $IoU\!=\!0.5$
and $85.18$/$80.20$/$79.14$ at $IoU\!=\!0.7$
(Table~\ref{tab:my-table-AP}), outperforming all prior SNN baselines
at $IoU\!=\!0.5$. The $1$--$3$ AP gap relative to vmem quantifies
the representational cost of the binary output transition, and is
visually evident in Fig.~\ref{fig:qualitative_comp}. Together, the
two variants bound the operating range of the proposed architecture:
vmem delivers peak accuracy with minimal hardware assumptions, while
spike prioritises full binary compatibility at a modest and
well-characterised accuracy cost, without any continuous intermediate
activations.

\subsection{Detection Performance}
Table~\ref{tab:my-table-AP} compares our model against CNN-based and
SNN-based detectors. CNN results are included for context. Our primary
claim is not to surpass full-precision CNNs in raw accuracy, but to
demonstrate that SNNs can approach their performance while delivering
substantial energy efficiency gains, as quantified in
Section~\ref{subsec:energy}.

Among SNN methods, vmem~(ours) achieves highly competitive performance
across all reported methods, surpassing SpikiLi~\cite{li2023spikeli} by
$+7.68$ Moderate and $+7.73$ Hard AP at $IoU\!=\!0.5$. At $IoU\!=\!0.7$,
the vmem variant establishes a new state-of-the-art among SNN methods on
all three difficulty tiers, achieving $90.11$ Easy, $85.50$ Moderate, and
$82.80$ Hard AP, surpassing SpikeCloudNet~\cite{ren2022spiking}. The fully
binary spike variant also demonstrates strong performance, outperforming
DeepSCNN~\cite{cao2015spiking} and SpikiLi~\cite{li2023spikeli} across
all metrics at $IoU\!=\!0.7$ despite relying exclusively on binary outputs
rather than continuous readouts.

The performance gap between vmem and spike---approximately $3$--$4$ AP
points at $IoU\!=\!0.5$ and $4$--$5$ AP points at $IoU\!=\!0.7$ directly
quantifies the representational cost of the final continuous-to-binary
transition. This gap is visually evident in
Fig.~\ref{fig:qualitative_comp}, where the fully binary \textit{spike}
variant exhibits characteristic failure modes on distant, sparse objects
and occasional orientation misalignments compared to the tighter
localization of the continuous \textit{vmem} readout. Importantly, this
gap is substantially narrower than the typical SNN-to-CNN accuracy gap
reported in the prior literature~\cite{li2023spikeli}, which we attribute
to the proposed spike-domain loss functions: the two-point temporal BCE
and Dice loss for keypoint detection, and the population-coded regression
loss for bounding box dimensions. These losses supervise the network
directly in the binary spike domain without requiring continuous
intermediate activations, enabling gradient information to flow
effectively across the full temporal depth of the spiking encoder-decoder.

Relative to CNNs, vmem~(ours) achieves $92.05$ Easy AP at $IoU\!=\!0.5$,
exceeding all four CNN baselines on this metric. At $IoU\!=\!0.7$,
vmem exceeds BEVDetNet~\cite{mohapatra2021bevdetnet} by $+7.70$ Moderate
AP and approaches PointPillars~\cite{lang2019pointpillars}~($86.10$) and
PointRCNN~\cite{shi2019pointrcnn}~($87.39$) on the Moderate tier, despite
operating with binary spike activations in all intermediate layers. This
result suggests that the primary accuracy bottleneck in our architecture
is the binary output readout rather than the spiking feature extractor,
motivating future work on improved spike-to-prediction decoding at the
output stage.

\subsection{Energy Efficiency Analysis}
\label{subsec:energy}

\subsubsection{\textbf{MAC Count Derivation}}
For a convolutional layer with $C_{\mathrm{in}}$ input channels,
$C_{\mathrm{out}}$ output channels, kernel size $K \times K$, and
output spatial resolution $H_{\mathrm{out}} \times W_{\mathrm{out}}$,
the MAC count is:
\begin{equation}
    N_{\mathrm{MAC}} = C_{\mathrm{in}} \times K^2
                       \times C_{\mathrm{out}}
                       \times H_{\mathrm{out}} \times W_{\mathrm{out}}.
    \label{eq:mac_count}
\end{equation}
Bias additions are excluded following standard convention, as their
contribution is negligible for all $C_{\mathrm{in}} \geq 2$, $K \geq 3$.

Each downsampling block~(DB) in the encoder comprises three convolutional
layers: a $5 \times 5$ feature extraction convolution (\texttt{c1}), a
$3 \times 3$ refinement convolution (\texttt{c2}), and a strided
$3 \times 3$ pooling convolution (\texttt{pool}) that halves the spatial
resolution. The skip connection concatenates the \texttt{c2} output with
the block input along the channel dimension before the pooling convolution.
We verify the MAC count for DB1 with input $16 \times 320 \times 320$
(the stem output):

\paragraph{\texttt{c1}: Conv2d($16 \to 32$, $5\!\times\!5$) at
$320\!\times\!320$}
\begin{equation}
    N_{\mathrm{MAC}}^{\texttt{c1}} =
    16 \times 5^2 \times 32 \times 320 \times 320
    = 1{,}310{,}720{,}000.
    \label{eq:db1_c1}
\end{equation}

\paragraph{\texttt{c2}: Conv2d($32 \to 32$, $3\!\times\!3$) at
$320\!\times\!320$}
\begin{equation}
    N_{\mathrm{MAC}}^{\texttt{c2}} =
    32 \times 3^2 \times 32 \times 320 \times 320
    = 943{,}718{,}400.
    \label{eq:db1_c2}
\end{equation}
The output of \texttt{c2} is concatenated with the $16$-channel block
input, yielding a $48$-channel feature map $(32 + 16 = 48)$ passed to
the pooling convolution.

\paragraph{\texttt{pool}: Conv2d($48 \to 48$, $3\!\times\!3$,
stride$\,{=}\,2$) at $160\!\times\!160$}
\begin{equation}
    N_{\mathrm{MAC}}^{\texttt{pool}} =
    48 \times 3^2 \times 48 \times 160 \times 160
    = 530{,}841{,}600.
    \label{eq:db1_pool}
\end{equation}
Summing all three layers gives the aggregate MAC count for DB1:
\begin{align}
    N_{\mathrm{MAC}}^{\mathrm{DB1}}
        &= N_{\mathrm{MAC}}^{\texttt{c1}}
         + N_{\mathrm{MAC}}^{\texttt{c2}}
         + N_{\mathrm{MAC}}^{\texttt{pool}} \notag \\
        &= 1{,}310{,}720{,}000
         + 943{,}718{,}400
         + 530{,}841{,}600 \notag \\
        &= 2{,}785{,}280{,}000,
    \label{eq:db1_total}
\end{align}
exactly matching the value reported in Table~\ref{tab:energy-detailed}.
The same procedure applied to DB2--DB4, the decoder blocks, and the
output heads recovers all remaining entries in the table, confirming
the internal consistency of the MAC estimates. The channel widths expand
progressively at each encoder stage: DB1 outputs 48 channels, DB2
outputs 112, DB3 outputs 240, and DB4 outputs 496, reflecting the
progressive feature expansion of the U-Net encoder.

\subsubsection{\textbf{From Multiply-Accumulate to Accumulate-Only Operations}}

The fundamental energy advantage of spiking networks over conventional
CNNs originates at the level of the individual synaptic operation.
In a CNN, every connection between two feature maps requires a
\textit{multiply-accumulate}~(MAC) operation: the presynaptic activation
(a continuous floating-point value) is multiplied by the synaptic weight
before being summed into the postsynaptic neuron. In an SNN, presynaptic
activations are binary spikes $s \in \{0,1\}$. A spike value of zero
contributes nothing to the postsynaptic neuron and requires no
computation. A spike value of one simply \textit{accumulates} the weight
directly, with no multiplication required. Every MAC is therefore
replaced by a cheaper \textit{accumulate-only}~(AC) operation.

Using 45\,nm technology constants~\cite{horowitz20141}---the
established reference node in neuromorphic energy
analysis~\cite{merolla2014million, roy2019towards}---this distinction
carries a concrete energy cost:
$E_{\mathrm{MAC}}\!=\!4.6$\,pJ versus $E_{\mathrm{AC}}\!=\!0.9$\,pJ---a
\textbf{5.1$\times$ reduction per active synaptic event}. This saving is
not an approximation or a systems-level engineering trick; it is a direct
consequence of eliminating the multiplier circuit from the datapath, and
applies identically on digital neuromorphic hardware~\cite{merolla2014million},
mixed-signal platforms~\cite{schemmel2010wafer}, and estimated CMOS
implementations~\cite{roy2019towards}.

\subsubsection{\textbf{Energy Model}}

For a layer with $N_{\mathrm{MAC}}$ synaptic connections, CNN inference
energy is:

\begin{equation}
    E_{\mathrm{CNN}} = N_{\mathrm{MAC}} \cdot E_{\mathrm{MAC}}.
    \label{eq:cnn_energy}
\end{equation}

In the SNN, synaptic events occur only when a presynaptic neuron fires.
Over $T$ timesteps with mean firing rate $\overline{r}$, the expected
number of active synaptic events is $N_{\mathrm{MAC}} \cdot \overline{r}
\cdot T$, each costing $E_{\mathrm{AC}}$:

\begin{equation}
    E_{\mathrm{SNN}} = N_{\mathrm{MAC}} \cdot \overline{r} \cdot T
                       \cdot E_{\mathrm{AC}}.
    \label{eq:snn_energy}
\end{equation}

The energy ratio between the two is:

\begin{equation}
    \frac{E_{\mathrm{CNN}}}{E_{\mathrm{SNN}}}
    = \frac{E_{\mathrm{MAC}}}{\overline{r} \cdot T \cdot E_{\mathrm{AC}}}
    = \frac{4.6}{\overline{r} \times 13 \times 0.9}.
    \label{eq:ratio}
\end{equation}

\noindent Equation~\ref{eq:ratio} reveals two independent levers for
energy reduction: the fixed $5.1\times$ MAC-to-AC gain, and a
further $1/(\overline{r} \cdot T)$ gain from spike sparsity. These
multiply together, so even moderate sparsity yields substantial savings.
At the network-wide mean firing rate of $\overline{r}\!=\!0.1181$,
Eq.~\ref{eq:ratio} predicts a ratio of
$4.6\,/\,(0.1181 \times 13 \times 0.9) = 3.33\times$, exactly
matching the measured total in Table~\ref{tab:energy-detailed}.

As a concrete block-level example, consider encoder block DB4 with
$N_{\mathrm{MAC}}\!=\!4{,}286{,}976{,}000$ and
$\overline{r}\!=\!0.0907$:

\begin{align}
    E_{\mathrm{CNN}}^{\mathrm{DB4}}
        &= 4{,}286{,}976{,}000 \times 4.6\,\mathrm{pJ}
         = 19{,}720\,\mu\mathrm{J}, \label{eq:db4_cnn} \\
    E_{\mathrm{SNN}}^{\mathrm{DB4}}
        &= 4{,}286{,}976{,}000 \times 0.0907 \times 13
           \times 0.9\,\mathrm{pJ}
         = 4{,}551\,\mu\mathrm{J}, \label{eq:db4_snn}
\end{align}

\noindent a $4.33\times$ reduction, driven by the MAC-to-AC
substitution combined with DB4's low firing rate of 9.07\%.

\subsubsection{\textbf{Network-Wide Results}}

Table~\ref{tab:energy-detailed} reports block-wise energy for all stages at
input resolution $320 \times 320$ with $T\!=\!13$ timesteps.
Total SNN inference energy is \textbf{95,170\,$\mu$J} versus
\textbf{316,935\,$\mu$J} for the equivalent CNN---an overall
\textbf{3.33$\times$ reduction}, meaning the SNN consumes less than
one third of the energy at identical task complexity and resolution.
It is important to note that this estimate is conservative: it is
derived from loop-based simulation in which the input is presented
$T\!=\!13$ times, counting energy for each pass. On dedicated
neuromorphic hardware, the $T$ timesteps unfold naturally along the
temporal dimension of the input stream, eliminating the loop overhead
entirely. In that setting the effective energy advantage scales to
$3.33 \times 13 \approx 43\times$ relative to a single-pass CNN.

The network exhibits a mean firing sparsity of \textbf{88.19\%}
($\overline{r}\!=\!0.1181$): on average, fewer than 1 in 8 neurons
fires at any given timestep.

\begin{figure}[!t]
    \centering
    \includegraphics[width=\columnwidth]{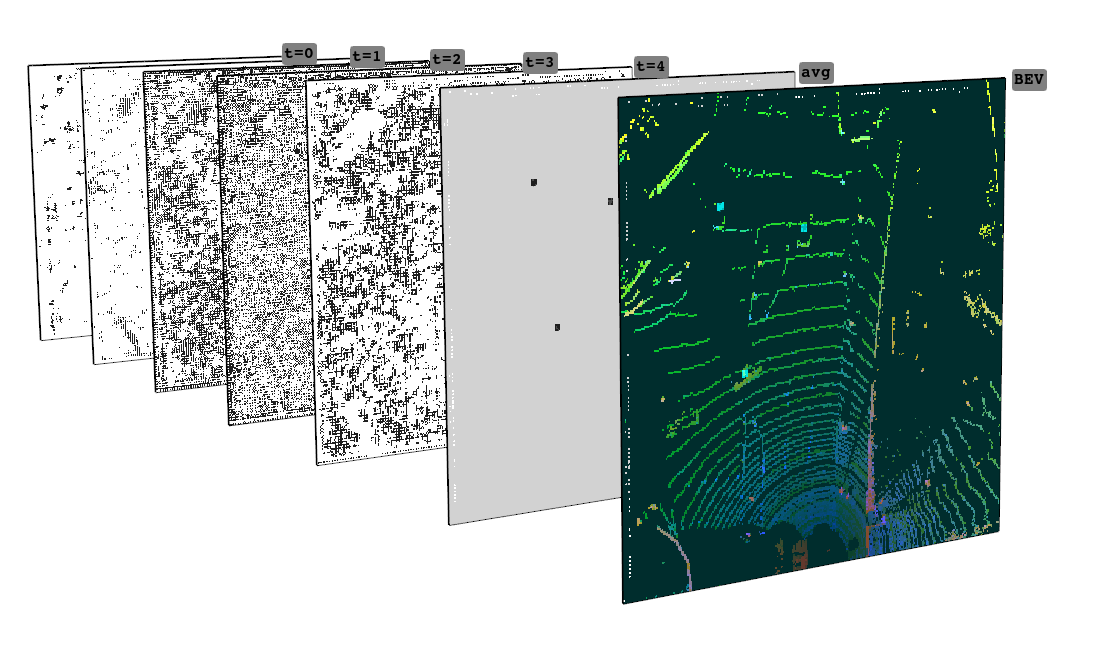}
    \caption{%
        \textbf{Spike activity over time for 5 time steps and spike rate for
        the keypoint detection head}: Each plane shows the binary spike
        pattern at a single timestep ($t=0$ through $t=4$), followed
        by the temporal mean firing rate (\textit{avg}) and the
        bird's-eye-view LiDAR input (\textit{BEV}). Cyan spots on the
        BEV highlight spatial regions with elevated average spike
        activity.
    }
    \label{fig:spike_activity_kp}
\end{figure}

Fig.~\ref{fig:spike_activity_kp} illustrates this sparsity
qualitatively for the keypoint detection head. The binary spike
maps at individual timesteps are highly sparse, with activity
concentrating progressively around object-center locations as
evidence accumulates across the simulation window. Notably, even
with only $T\!=\!5$ timesteps shown, the temporal mean firing rate
map (\textit{avg}) already exhibits clearly elevated activity at
vehicle locations in the bird's eye view input (\textit{BEV})---demonstrating
that the network converges to a meaningful spatial representation
rapidly, well within the full $T\!=\!13$ timestep budget.

\begin{figure}[!t]
    \centering
    \includegraphics[width=\columnwidth]{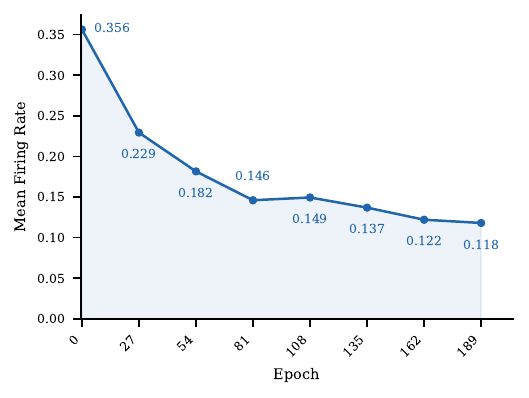}
    \caption{\textbf{Mean firing rate of the proposed SNN across training
             epochs}: The network learns progressively sparser spike
             representations.}
    \label{fig:firing_rate}
\end{figure}

As shown in Fig.~\ref{fig:firing_rate}, this sparsity is not a
fixed property of the architecture but an \textit{emergent} result
of training: the mean firing rate decreases monotonically from
$0.356$ at epoch~0 to $0.118$ at epoch~189, indicating that the
network progressively learns sparser spike representations as
it converges. Since inactive neurons contribute exactly
zero synaptic operations, this sparsity acts as a second independent
multiplier on the energy savings already provided by the
MAC$\,{\to}\,$AC substitution. Together, these two mechanisms account
for the full $3.33\times$ gain.

The decoder dominates SNN energy at 78.37\%, driven by the large MAC
counts of the transposed convolution upsampling blocks operating at
progressively higher spatial resolutions. UB3\_4 alone accounts for
28.64\% of total SNN energy, yet still achieves $2.78\times$ over its
CNN equivalent. The encoder attains $4.88\times$ overall, with DB3
reaching $5.77\times$ as the most efficient encoder block. The three
output heads together account for only 6.96\% of total SNN
energy---confirming that the multi-task prediction overhead (keypoint,
bounding box, and rotation heads) is negligible relative to the
backbone, an important property for scaling to additional detection tasks.

\subsection{Limitations}

Consistent with~\cite{mohapatra2021bevdetnet}, heading direction is not
predicted, following the BEV benchmarking convention that exploits
orientation symmetry. The self-coding strategy does not exploit temporal
variation between consecutive LiDAR scans. In a deployed streaming system
this limitation is naturally resolved, but multi-frame training to
leverage motion cues remains an avenue for future work. On-chip profiling
of the spike variant on physical neuromorphic hardware would further
validate the energy estimates reported here.


\section{Conclusion}
\label{sec:conclusion}
We presented an end-to-end spiking encoder-decoder network for
LiDAR bird's eye view 3D object detection trained with surrogate
gradients, offering two inference variants: a membrane potential
variant~(vmem) for maximum accuracy and a fully binary spiking
variant~(spike) for direct neuromorphic deployment. The vmem variant
achieves $92.05$/$87.04$/$86.51$ AP at $IoU\!=\!0.5$ and
$90.11$/$85.50$/$82.80$ at $IoU\!=\!0.7$, establishing a new
state-of-the-art among SNN-based detectors and approaching
full-precision CNN baselines. The spike variant achieves
$88.51$/$85.72$/$83.14$ AP at $IoU\!=\!0.5$ using exclusively
binary activations, outperforming all prior SNN-based BEV detectors
at this threshold.

Two novel spike-domain loss functions were central to these results:
a two-point temporal BCE and masked Dice loss for keypoint detection
operating directly on heatmap targets, and a population-coded
regression loss for bounding box estimation that eliminates the need
to discretise continuous targets. Together, these losses enable
gradient information to flow effectively across the full temporal
depth of the spiking encoder-decoder without requiring continuous
intermediate activations.

A block-wise energy analysis at $320 \times 320$ input resolution
confirms a $3.33\times$ reduction in synaptic operation energy over
an equivalent CNN, driven by the MAC$\,{\to}\,$AC operation
substitution and $88.19\%$ network-wide firing sparsity, both
emerging naturally from training without explicit regularisation.
This estimate is conservative: on dedicated neuromorphic hardware
where $T$ timesteps unfold naturally along the input stream rather
than through explicit simulation loops, the effective energy
advantage scales to approximately $43\times$.

The primary avenue for future work is realising true event-driven
inference by developing sparse convolution primitives that restrict
computation to the spatial neighbourhoods of active spike events,
converting the theoretical MAC$\,{\to}\,$AC saving into a measured
reduction in runtime and on-chip energy. Additional directions
include multi-frame training to exploit temporal variation between
consecutive LiDAR scans, on-chip profiling to validate the energy
estimates against measured silicon power, and a study of the
accuracy--energy trade-off as a function of the simulation timestep
budget $T$.

\bibliographystyle{IEEEtran}
\bibliography{bib/references}

@article{eshraghian2023training,
  title={Training spiking neural networks using lessons from deep learning},
  author={Eshraghian, Jason K and Ward, Max and Neftci, Emre O and Wang, Xinxin and Lenz, Gregor and Dwivedi, Girish and Bennamoun, Mohammed and Jeong, Doo Seok and Lu, Wei D},
  journal={Proceedings of the IEEE},
  volume={111},
  number={9},
  pages={1016--1054},
  year={2023},
  publisher={IEEE}
}

@article{geiger2013vision,
  title={Vision meets robotics: The kitti dataset},
  author={Geiger, Andreas and Lenz, Philip and Stiller, Christoph and Urtasun, Raquel},
  journal={The international journal of robotics research},
  volume={32},
  number={11},
  pages={1231--1237},
  year={2013},
  publisher={Sage Publications Sage UK: London, England}
}

@inproceedings{engelcke2017vote3deep,
  title={Vote3deep: Fast object detection in 3d point clouds using efficient convolutional neural networks},
  author={Engelcke, Martin and Rao, Dushyant and Wang, Dominic Zeng and Tong, Chi Hay and Posner, Ingmar},
  booktitle={2017 IEEE International Conference on Robotics and Automation (ICRA)},
  pages={1355--1361},
  year={2017},
  organization={IEEE}
}

@inproceedings{zhou2018voxelnet,
  title={Voxelnet: End-to-end learning for point cloud based 3d object detection},
  author={Zhou, Yin and Tuzel, Oncel},
  booktitle={Proceedings of the IEEE conference on computer vision and pattern recognition},
  pages={4490--4499},
  year={2018}
}

@article{yan2018second,
  title={Second: Sparsely embedded convolutional detection},
  author={Yan, Yan and Mao, Yuxing and Li, Bo},
  journal={Sensors},
  volume={18},
  number={10},
  pages={3337},
  year={2018},
  publisher={Multidisciplinary Digital Publishing Institute}
}

@inproceedings{shi2019pointrcnn,
  title={Pointrcnn: 3d object proposal generation and detection from point cloud},
  author={Shi, Shaoshuai and Wang, Xiaogang and Li, Hongsheng},
  booktitle={Proceedings of the IEEE/CVF conference on computer vision and pattern recognition},
  pages={770--779},
  year={2019}
}

@article{qi2017pointnet++,
  title={Pointnet++: Deep hierarchical feature learning on point sets in a metric space},
  author={Qi, Charles Ruizhongtai and Yi, Li and Su, Hao and Guibas, Leonidas J},
  journal={Advances in neural information processing systems},
  volume={30},
  year={2017}
}

@inproceedings{shi2020pv,
  title={Pv-rcnn: Point-voxel feature set abstraction for 3d object detection},
  author={Shi, Shaoshuai and Guo, Chaoxu and Jiang, Li and Wang, Zhe and Shi, Jianping and Wang, Xiaogang and Li, Hongsheng},
  booktitle={Proceedings of the IEEE/CVF conference on computer vision and pattern recognition},
  pages={10529--10538},
  year={2020}
}

@inproceedings{lang2019pointpillars,
  title={Pointpillars: Fast encoders for object detection from point clouds},
  author={Lang, Alex H and Vora, Sourabh and Caesar, Holger and Zhou, Lubing and Yang, Jiong and Beijbom, Oscar},
  booktitle={Proceedings of the IEEE/CVF conference on computer vision and pattern recognition},
  pages={12697--12705},
  year={2019}
}

@inproceedings{qi2017pointnet,
  title={Pointnet: Deep learning on point sets for 3d classification and segmentation},
  author={Qi, Charles R and Su, Hao and Mo, Kaichun and Guibas, Leonidas J},
  booktitle={Proceedings of the IEEE conference on computer vision and pattern recognition},
  pages={652--660},
  year={2017}
}

@inproceedings{mohapatra2021bevdetnet,
  title={BEVDetNet: Bird's eye view LiDAR point cloud based real-time 3D object detection for autonomous driving},
  author={Mohapatra, Sambit and Yogamani, Senthil and Gotzig, Heinrich and Milz, Stefan and Mader, Patrick},
  booktitle={2021 IEEE International Intelligent Transportation Systems Conference (ITSC)},
  pages={2809--2815},
  year={2021},
  organization={IEEE}
}

@inproceedings{lis2025lift,
  title={LiFT: Lightweight, FPGA-tailored 3D object detection based on LiDAR data},
  author={Lis, Konrad and Kryjak, Tomasz and Gorgo{\'n}, Marek},
  booktitle={International Workshop on Design and Architectures for Signal and Image Processing},
  pages={28--40},
  year={2025},
  organization={Springer}
}

@article{naich2024lidar,
  title={LiDAR-based intensity-aware outdoor 3D object detection},
  author={Naich, Ammar Yasir and Carri{\'o}n, Jes{\'u}s Requena},
  journal={Sensors},
  volume={24},
  number={9},
  pages={2942},
  year={2024},
  publisher={MDPI}
}

@article{neftci2019surrogate,
  title={Surrogate gradient learning in spiking neural networks: Bringing the power of gradient-based optimization to spiking neural networks},
  author={Neftci, Emre O and Mostafa, Hesham and Zenke, Friedemann},
  journal={IEEE Signal Processing Magazine},
  volume={36},
  number={6},
  pages={51--63},
  year={2019},
  publisher={IEEE}
}

@article{cao2015spiking,
  title={Spiking deep convolutional neural networks for energy-efficient object recognition},
  author={Cao, Yongqiang and Chen, Yang and Khosla, Deepak},
  journal={International Journal of Computer Vision},
  volume={113},
  number={1},
  pages={54--66},
  year={2015},
  publisher={Springer}
}

@article{rueckauer2017conversion,
  title={Conversion of continuous-valued deep networks to efficient event-driven networks for image classification},
  author={Rueckauer, Bodo and Lungu, Iulia-Alexandra and Hu, Yuhuang and Pfeiffer, Michael and Liu, Shih-Chii},
  journal={Frontiers in neuroscience},
  volume={11},
  pages={682},
  year={2017},
  publisher={Frontiers Media SA}
}

@inproceedings{deng2009imagenet,
  title={Imagenet: A large-scale hierarchical image database},
  author={Deng, Jia and Dong, Wei and Socher, Richard and Li, Li-Jia and Li, Kai and Fei-Fei, Li},
  booktitle={2009 IEEE conference on computer vision and pattern recognition},
  pages={248--255},
  year={2009},
  organization={Ieee}
}

@techreport{krizhevsky2009learning,
  title={Learning multiple layers of features from tiny images},
  author={Krizhevsky, Alex and Hinton, Geoffrey and others},
  year={2009},
  institution={University of Toronto},
  publisher={Toronto, ON, Canada}
}

@inproceedings{diehl2015fast,
  title={Fast-classifying, high-accuracy spiking deep networks through weight and threshold balancing},
  author={Diehl, Peter U and Neil, Daniel and Binas, Jonathan and Cook, Matthew and Liu, Shih-Chii and Pfeiffer, Michael},
  booktitle={2015 International joint conference on neural networks (IJCNN)},
  pages={1--8},
  year={2015},
  organization={ieee}
}

@inproceedings{kim2020spiking,
  title={Spiking-yolo: spiking neural network for energy-efficient object detection},
  author={Kim, Seijoon and Park, Seongsik and Na, Byunggook and Yoon, Sungroh},
  booktitle={Proceedings of the AAAI conference on artificial intelligence},
  volume={34},
  number={07},
  pages={11270--11277},
  year={2020}
}

@article{qu2024spiking,
  title={Spiking neural network for ultralow-latency and high-accurate object detection},
  author={Qu, Jinye and Gao, Zeyu and Zhang, Tielin and Lu, Yanfeng and Tang, Huajin and Qiao, Hong},
  journal={IEEE Transactions on Neural Networks and Learning Systems},
  volume={36},
  number={3},
  pages={4934--4946},
  year={2024},
  publisher={IEEE}
}

@article{wu2018spatio,
  title={Spatio-temporal backpropagation for training high-performance spiking neural networks},
  author={Wu, Yujie and Deng, Lei and Li, Guoqi and Zhu, Jun and Shi, Luping},
  journal={Frontiers in neuroscience},
  volume={12},
  pages={331},
  year={2018}
}

@article{tao2024spiking,
  title={Spiking PointCNN: An Efficient Converted Spiking Neural Network under a Flexible Framework},
  author={Tao, Yingzhi and Wu, Qiaoyun},
  journal={Electronics},
  volume={13},
  number={18},
  pages={3626},
  year={2024},
  publisher={MDPI},
  doi={10.3390/electronics13183626}
}

@inproceedings{mohapatra2022spikili,
  title={SpikiLi: A Spiking Simulation of LiDAR based Real-time Object Detection for Autonomous Driving},
  author={Mohapatra, Sambit and Mesquida, Thomas and Hodaei, Maryam and Yogamani, Senthil and Gotzig, Heinrich and M{\"a}der, Patrick},
  booktitle={2022 IEEE Intelligent Vehicles Symposium (IV)},
  pages={1118--1125},
  year={2022},
  organization={IEEE},
  doi={10.1109/IV51971.2022.9827051}
}

@inproceedings{li2018pointcnn,
  title={PointCNN: Convolution On X-Transformed Points},
  author={Li, Yangyan and Bu, Rui and Sun, Mingchao and Wu, Wei and Di, Xinhan and Chen, Baoquan},
  booktitle={Advances in Neural Information Processing Systems (NeurIPS)},
  volume={31},
  pages={820--830},
  year={2018}
}

@inproceedings{lan2023efficient,
  title={Efficient Converted Spiking Neural Network for 3D and 2D Classification},
  author={Lan, Shukai and Zhang, Mi and Wu, Qiaoyun and others},
  booktitle={Proceedings of the IEEE/CVF International Conference on Computer Vision (ICCV)},
  pages={10356--10365},
  year={2023}
}

@inproceedings{li2025brain,
  title={Brain-Inspired Spiking Neural Networks for Energy-Efficient Object Detection},
  author={Li, Zhenglong and Yao, Man and Qiu, Xuerui and others},
  booktitle={Proceedings of the IEEE/CVF Conference on Computer Vision and Pattern Recognition (CVPR)},
  year={2025}
}

@article{adrian1926impulses,
  title={The impulses produced by sensory nerve endings},
  author={Adrian, Edgar Douglas and Zotterman, Yngve},
  journal={The Journal of physiology},
  volume={61},
  number={4},
  pages={465--483},
  year={1926},
  publisher={Wiley}
}

@article{maass1997networks,
  title={Networks of spiking neurons: the third generation of neural network models},
  author={Maass, Wolfgang},
  journal={Neural networks},
  volume={10},
  number={9},
  pages={1659--1671},
  year={1997},
  publisher={Elsevier}
}

@article{thorpe1996rapid,
  title={Rapid visual processing using spike asynchrony},
  author={Thorpe, Simon and Gautrais, Jacques},
  journal={Advances in neural information processing systems},
  volume={9},
  year={1996}
}

@article{gallego2020event,
  title={Event-based vision: A survey},
  author={Gallego, Guillermo and Delbr{\"u}ck, Tobi and Orchard, Garrick and Bartolozzi, Chiara and Taba, Brian and Censi, Andrea and Leutenegger, Stefan and Davison, Andrew J and Conradt, J{\"o}rg and Daniilidis, Kostas and others},
  journal={IEEE transactions on pattern analysis and machine intelligence},
  volume={44},
  number={1},
  pages={154--180},
  year={2020},
  publisher={IEEE}
}

@inproceedings{su2023deep,
  title={Deep directly-trained spiking neural networks for object detection},
  author={Su, Qiaoyi and Chou, Yuhong and Hu, Yifan and Li, Jianing and Mei, Shijie and Zhang, Ziyang and Li, Guoqi},
  booktitle={Proceedings of the IEEE/CVF International Conference on Computer Vision},
  pages={6555--6565},
  year={2023}
}

@inproceedings{ronneberger2015u,
  title={U-net: Convolutional networks for biomedical image segmentation},
  author={Ronneberger, Olaf and Fischer, Philipp and Brox, Thomas},
  booktitle={International Conference on Medical image computing and computer-assisted intervention},
  pages={234--241},
  year={2015},
  organization={Springer}
}

@article{mohapatra2025lidar,
  title={Lidar-bevmtn: Real-time lidar bird’s-eye view multi-task perception network for autonomous driving},
  author={Mohapatra, Sambit and Yogamani, Senthil and Kumar, Varun Ravi and Milz, Stefan and Gotzig, Heinrich and M{\"a}der, Patrick},
  journal={IEEE transactions on intelligent transportation systems},
  volume={26},
  number={2},
  pages={1547--1561},
  year={2025},
  publisher={IEEE}
}

@inproceedings{mohapatra2021limoseg,
	title        = {{LiMoSeg: Real-time Bird's Eye View based LiDAR Motion Segmentation}},
	author       = {Mohapatra, Sambit and Hodaei, Mona and Yogamani, Senthil and Milz, Stefan and others},
year={2022},
booktitle={Proceedings of the 17th International Joint Conference on Computer Vision, Imaging and Computer Graphics Theory and Applications (VISIGRAPP 2022) }
}

@inproceedings{barrera2020birdnet+,
  title={Birdnet+: End-to-end 3d object detection in lidar bird’s eye view},
  author={Barrera, Alejandro and Guindel, Carlos and Beltr{\'a}n, Jorge and Garc{\'\i}a, Fernando},
  booktitle={2020 IEEE 23rd International Conference on Intelligent Transportation Systems (ITSC)},
  pages={1--6},
  year={2020},
  organization={IEEE}
}

@article{pouget2000,
  author    = {Pouget, Alexandre and Dayan, Peter and Zemel, Richard},
  title     = {Information Processing with Population Codes},
  journal   = {Nature Reviews Neuroscience},
  volume    = {1},
  number    = {2},
  pages     = {125--132},
  year      = {2000},
  publisher = {Nature Publishing Group},
  doi       = {10.1038/35036228}
}

@article{neftci2019,
  author    = {Neftci, Emre O. and Mostafa, Hesham and Zenke, Friedemann},
  title     = {Surrogate Gradient Learning in Spiking Neural Networks:
               Bringing the Power of Gradient-Based Optimization
               to Spiking Neural Networks},
  journal   = {IEEE Signal Processing Magazine},
  volume    = {36},
  number    = {6},
  pages     = {51--63},
  year      = {2019},
  doi       = {10.1109/MSP.2019.2931595}
}

@inproceedings{lee2016,
  author    = {Lee, Jun Haeng and Delbruck, Tobi and Pfeiffer, Michael},
  title     = {Training Deep Spiking Neural Networks Using Backpropagation},
  booktitle = {Frontiers in Neuroscience},
  volume    = {10},
  pages     = {508},
  year      = {2016},
  doi       = {10.3389/fnins.2016.00508}
}

@inproceedings{chen2015multi,
  author    = {Chen, Xiaozhi and Kundu, Kaustav and Zhang, Ziyu
               and Ma, Huimin and Fidler, Sanja and Urtasun, Raquel},
  title     = {Monocular 3{D} Object Detection for Autonomous Driving},
  booktitle = {Proceedings of the IEEE Conference on Computer Vision
               and Pattern Recognition (CVPR)},
  pages     = {2147--2156},
  year      = {2016}
}

@article{kingma2014adam,
  author    = {Kingma, Diederik P. and Ba, Jimmy},
  title     = {Adam: A Method for Stochastic Optimization},
  journal   = {arXiv preprint arXiv:1412.6980},
  year      = {2014}
}

@inproceedings{loshchilov2016sgdr,
  author    = {Loshchilov, Ilya and Hutter, Frank},
  title     = {{SGDR}: Stochastic Gradient Descent with Warm Restarts},
  booktitle = {International Conference on Learning Representations (ICLR)},
  year      = {2017}
}

@inproceedings{caesar2020nuscenes,
  author    = {Caesar, Holger and Bankiti, Varun and Lang, Alex H.
               and Vora, Sourabh and Liong, Venice Erin and Xu, Qiang
               and Krishnan, Anush and Pan, Yu and Baldan, Giancarlo
               and Beijbom, Oscar},
  title     = {nu{S}cenes: A Multimodal Dataset for Autonomous Driving},
  booktitle = {Proceedings of the IEEE Conference on Computer Vision
               and Pattern Recognition (CVPR)},
  pages     = {11621--11631},
  year      = {2020}
}

@article{auge2021survey,
  author    = {Auge, Daniel and Hille, Julian and Mueller, Etienne
               and Knoll, Alois},
  title     = {A Survey of Encoding Techniques for Signal Processing
               in Spiking Neural Networks},
  journal   = {Neural Processing Letters},
  volume    = {53},
  pages     = {4395--4429},
  year      = {2021},
  doi       = {10.1007/s11063-021-10562-2}
}

@article{roy2019towards,
  author    = {Roy, Kaushik and Jaiswal, Akhilesh and Panda, Priyadarshini},
  title     = {Towards Spike-Based Machine Intelligence with Neuromorphic
               Computing},
  journal   = {Nature},
  volume    = {575},
  pages     = {607--617},
  year      = {2019},
  doi       = {10.1038/s41586-019-1677-2}
}

@article{merolla2014million,
  author    = {Merolla, Paul A. and Arthur, John V. and Alvarez-Icaza,
               Rodrigo and Cassidy, Andrew S. and Sawada, Jun and
               Akopyan, Filipp and Jackson, Bryan L. and Imam, Nabil
               and Guo, Chen and Nakamura, Yutaka and others},
  title     = {A Million Spiking-Neuron Integrated Circuit with a
               Scalable Communication Network and Interface},
  journal   = {Science},
  volume    = {345},
  number    = {6197},
  pages     = {668--673},
  year      = {2014},
  doi       = {10.1126/science.1254642}
}

@inproceedings{orchard2021loihi,
  author    = {Orchard, Garrick and Frady, E. Paxon and Rubin, Daniel B.
               and Sanborn, Sophia and Shrestha, Sumit Bam and
               Sommer, Friedrich T. and Davies, Mike},
  title     = {Efficient Neuromorphic Signal Processing with {Loihi} 2},
  booktitle = {IEEE Workshop on Signal Processing Systems (SiPS)},
  pages     = {254--259},
  year      = {2021},
  doi       = {10.1109/SiPS52927.2021.00053}
}

@inproceedings{furber2014spinnaker,
  author    = {Furber, Steve B. and Galluppi, Francesco and Temple, Steve
               and Plana, Luis A.},
  title     = {The {SpiNNaker} Project},
  journal   = {Proceedings of the IEEE},
  volume    = {102},
  number    = {5},
  pages     = {652--665},
  year      = {2014},
  doi       = {10.1109/JPROC.2014.2304638}
}

@inproceedings{schemmel2010wafer,
  author    = {Schemmel, Johannes and Br\"{u}derle, Daniel and
               Gr\"{u}bl, Andreas and Hock, Matthias and Meier, Karlheinz
               and Millner, Sebastian},
  title     = {A Wafer-Scale Neuromorphic Hardware System for Large-Scale
               Neural Modeling},
  booktitle = {Proceedings of the IEEE International Symposium on
               Circuits and Systems (ISCAS)},
  pages     = {1947--1950},
  year      = {2010},
  doi       = {10.1109/ISCAS.2010.5536970}
}

@inproceedings{horowitz20141,
  author    = {Horowitz, Mark},
  title     = {1.1 Computing's Energy Problem (and What We Can Do About It)},
  booktitle = {IEEE International Solid-State Circuits Conference (ISSCC)},
  pages     = {10--14},
  year      = {2014},
  doi       = {10.1109/ISSCC.2014.6757323}
}

@inproceedings{yang2018pixor,
  author    = {Yang, Bin and Luo, Wenjie and Urtasun, Raquel},
  title     = {{PIXOR}: Real-Time {3D} Object Detection from Point Clouds},
  booktitle = {Proceedings of the IEEE Conference on Computer Vision
               and Pattern Recognition (CVPR)},
  pages     = {7652--7660},
  year      = {2018}
}

@inproceedings{ren2022spiking,
  author    = {Ren, Huajin and Zhao, Zhongjian and Lombardini, Alberto
               and Tang, Guoqi and Li, Peng},
  title     = {Spiking {PointNet}: Spiking Neural Networks for Point
               Clouds},
  booktitle = {Advances in Neural Information Processing Systems
               (NeurIPS)},
  year      = {2023}
}

@inproceedings{li2023spikeli,
  author    = {Li, Xiang and Bhatt, Pranshu and Li, Rui and Zhang, Wei
               and Bhatt, Utkarsh},
  title     = {{SpikiLi}: A Spiking Based {LiDAR} Point Cloud Object
               Detection Model for Autonomous Driving},
  booktitle = {IEEE International Conference on Acoustics, Speech and
               Signal Processing (ICASSP)},
  year      = {2023}
}

@article{li2020lidar,
  author    = {Li, Yizhou and Ibanez-Guzman, Javier},
  title     = {{LiDAR} for Autonomous Driving: The
               Principles, Challenges, and Trends for
               Automotive {LiDAR} and Perception Systems},
  journal   = {IEEE Signal Processing Magazine},
  volume    = {37},
  number    = {4},
  pages     = {50--61},
  year      = {2020},
  doi       = {10.1109/MSP.2020.2973402}
}

@inproceedings{milletari2016vnet,
  author    = {Milletari, Fausto and Navab, Nassir and
               Ahmadi, Seyed-Ahmad},
  title     = {V-{N}et: Fully Convolutional Neural Networks
               for Volumetric Medical Image Segmentation},
  booktitle = {International Conference on 3D Vision (3DV)},
  pages     = {565--571},
  year      = {2016},
  doi       = {10.1109/3DV.2016.79}
}

@inproceedings{sudre2017generalised,
  author    = {Sudre, Carole H. and Li, Wenqi and Vercauteren,
               Tom and Ourselin, S{\'e}bastien and
               Cardoso, M. Jorge},
  title     = {Generalised Dice Overlap as a Deep Learning
               Loss Function for Highly Unbalanced
               Segmentations},
  booktitle = {Deep Learning in Medical Image Analysis and
               Multimodal Learning for Clinical Decision
               Support (DLMIA)},
  pages     = {240--248},
  year      = {2017},
  doi       = {10.1007/978-3-319-67558-9\_28}
}

@inproceedings{szegedy2016rethinking,
  author    = {Szegedy, Christian and Vanhoucke, Vincent and
               Ioffe, Sergey and Shlens, Jon and Wojna, Zbigniew},
  title     = {Rethinking the Inception Architecture for
               Computer Vision},
  booktitle = {Proceedings of the IEEE Conference on Computer
               Vision and Pattern Recognition (CVPR)},
  pages     = {2818--2826},
  year      = {2016}
}

@article{pouget2000information,
  author    = {Pouget, Alexandre and Dayan, Peter and
               Zemel, Richard},
  title     = {Information Processing with Population Codes},
  journal   = {Nature Reviews Neuroscience},
  volume    = {1},
  pages     = {125--132},
  year      = {2000},
  doi       = {10.1038/35036228}
}

@article{lian2020deep,
  author    = {Lian, Shuiying and Luo, Jiamin and Zhao,
               Zhichao and Li, Shuai and Yu, Shaoying and
               Deng, Liang},
  title     = {Deep {SCNN}-Based Real-Time Object Detection
               for Self-Driving Vehicles Using {LiDAR}
               Temporal Data},
  journal   = {IEEE Access},
  volume    = {8},
  pages     = {76903--76912},
  year      = {2020},
  doi       = {10.1109/ACCESS.2020.2990416}
}

@inproceedings{ren2024spikingpointnet,
  author    = {Ren, Dayong and Ma, Zhe and Chen, Yuanpei and
               Peng, Weihang and Liu, Xiaode and Zhang, Yuhan
               and Guo, Yufei},
  title     = {Spiking {PointNet}: Spiking Neural Networks
               for Point Clouds},
  booktitle = {Advances in Neural Information Processing
               Systems (NeurIPS)},
  volume    = {36},
  year      = {2023}
}

@inproceedings{wu2024pointtospike,
  author    = {Wu, Qiaoyun and Zhang, Quanlu and Tan, Chao
               and Zhou, Yi and Sun, Chang},
  title     = {Point-to-Spike Residual Learning for
               Energy-Efficient {3D} Point Cloud
               Classification},
  booktitle = {Proceedings of the AAAI Conference on
               Artificial Intelligence},
  volume    = {38},
  pages     = {6092--6099},
  year      = {2024}
}

@inproceedings{cordone2022detection,
  author    = {Cordone, Lo\"{i}c and Miramond, Beno\^{i}t
               and Thierion, Philippe},
  title     = {Object Detection with Spiking Neural Networks
               on Automotive Event Data},
  booktitle = {International Joint Conference on Neural
               Networks (IJCNN)},
  pages     = {1--8},
  year      = {2022}
}

@inproceedings{zhu2024spiking,
  author    = {Zhu, Rui-Jie and Wang, Ziqing and Gilpin,
               Leilani and Eshraghian, Jason K.},
  title     = {Autonomous Driving with Spiking Neural
               Networks},
  booktitle = {Advances in Neural Information Processing
               Systems (NeurIPS)},
  year      = {2024}
}

@article{neumeier2025spikeclouds,
  title={SpikeClouds: Streaming Spike-Based Processing of LiDAR for Fast and Efficient Object Detection},
  author={Neumeier, Michael and Fasfous, Nael and Li, Bing and von Arnim, Axel},
  journal={IEEE Robotics and Automation Letters},
  year={2025},
  publisher={IEEE}
}

@inproceedings{law2018cornernet,
  title     = {CornerNet: Detecting Objects as Paired Keypoints},
  author    = {Law, Hei and Deng, Jia},
  booktitle = {Proceedings of the European Conference on Computer Vision (ECCV)},
  pages     = {734--750},
  year      = {2018}
}

@inproceedings{lin2017focal,
  title     = {Focal Loss for Dense Object Detection},
  author    = {Lin, Tsung-Yi and Goyal, Priya and Girshick, Ross and
               He, Kaiming and Doll{\'a}r, Piotr},
  booktitle = {Proceedings of the IEEE International Conference on
               Computer Vision (ICCV)},
  pages     = {2980--2988},
  year      = {2017}
}

@inproceedings{kumar2018near,
  title={Near-field depth estimation using monocular fisheye camera: A semi-supervised learning approach using sparse LiDAR data},
  author={Kumar, Varun Ravi and Milz, Stefan and Witt, Christian and Simon, Martin and Amende, Karl and Petzold, Johannes and Yogamani, Senthil and Pech, Timo},
  booktitle={CVPR Workshop},
  volume={7},
  pages={2},
  year={2018}
}

@conference{uricar2019challenges,
  title={Challenges in designing datasets and validation for autonomous driving},
  author={Uric{\'a}r, Michal and Hurych, David and Krizek, Pavel and Yogamani, Senthil},
booktitle={Proceedings of the International Joint Conference on Computer Vision, Imaging and Computer Graphics Theory and Applications (VISAPP)},
year={2019}
}

@inproceedings{schramm2024bevcar,
  title={Bevcar: Camera-radar fusion for bev map and object segmentation},
  author={Schramm, Jonas and V{\"o}disch, Niclas and Petek, K{\"u}rsat and Kiran, B Ravi and Yogamani, Senthil and Burgard, Wolfram and Valada, Abhinav},
  booktitle={2024 IEEE/RSJ International Conference on Intelligent Robots and Systems (IROS)},
  pages={1435--1442},
  year={2024},
  organization={IEEE}
}

@book{joseph2021autonomous,
  title={Autonomous driving and advanced driver-assistance systems (ADAS): applications, development, legal issues, and testing},
  author={Joseph, Lentin and Mondal, Amit Kumar},
  year={2021},
  publisher={CRC Press}
}

\end{document}